\documentclass[lettersize,journal]{IEEEtran}
\usepackage{amsmath,amsfonts}
\usepackage{bm}
\usepackage[numbers]{natbib}
\usepackage{algorithm}
\usepackage{algorithmic}
\usepackage{array}
\usepackage[caption=false,font=normalsize,labelfont=sf,textfont=sf]{subfig}
\usepackage{textcomp}
\usepackage{stfloats}
\usepackage{url}
\usepackage{verbatim}
\usepackage{graphicx}
\usepackage{cite}

\usepackage{amsmath}
\usepackage{amssymb}
\usepackage{mathtools}
\usepackage{hyperref} 
\usepackage{enumitem}
\usepackage{multirow}
\usepackage{relsize}
\usepackage{subcaption}
\usepackage{todonotes}
\usepackage{subfig}
\usepackage{booktabs} 
\usepackage{amsthm}
\newtheorem{theorem}{Theorem}

\usepackage[table]{xcolor}
\usepackage{threeparttable}
\usepackage{booktabs}   
\usepackage{balance} 
\usepackage{placeins}
\usepackage{ragged2e}

\newtheorem{proposition}[theorem]{Proposition}
\newtheorem{corollary}[theorem]{Corollary}

\newtheorem{remark}[theorem]{Remark}

\newcommand{\R}{\mathbb{R}}
\newcommand{\E}{\mathbb{E}}
\newcommand{\vect}[1]{\mathbf{#1}}
\newcommand{\mat}[1]{\mathbf{#1}}
\newcommand{\method}{UniFlow}
\begin{document}

\title{UniFlow: Unified Multi-Scale Flow Matching for Time Series Generation}

\author{Shibo Feng*, Wei Li*, Songhua Liu, Min Wu~\IEEEmembership{Senior Member, IEEE}, Chunyan Miao$^\dagger$~\IEEEmembership{Fellow, IEEE}, Peilin Zhao$^\dagger$~\IEEEmembership{Senior Member, IEEE}
\IEEEcompsocitemizethanks{
\IEEEcompsocthanksitem Shibo Feng and Chuanyan Miao are with the College of Computing and Data Science, Nanyang Technological University, Singapore.
Email:\{shibo001,ascymiao\}@ntu.edu.sg.
\IEEEcompsocthanksitem Wei Li is with Shanghai Jiao Tong University and Beijing Zhongguancun Academy, Shanghai, China. Email: liwei008009@163.com
\IEEEcompsocthanksitem Min Wu is with the Institute for Infocomm Research and AIMfg, A* STAR,
Singapore 138634. Email: wumin@i2r.a-star.edu.sg.
\IEEEcompsocthanksitem Peilin Zhao and Songhua Liu are with the School of Artificial Intelligence, Shanghai Jiao Tong University, Shanghai, China. Email: \{liusonghua, peilinzhao\}@sjtu.edu.cn.
\IEEEcompsocthanksitem * Authors contributed equally, $\dagger$ Corresponding author.
}
}


\markboth{Preprint}%
{Shell \MakeLowercase{\textit{et al.}}: A Sample Article Using IEEEtran.cls for IEEE Journals}


\maketitle

\begin{abstract}
        Real-world time series exhibit complex dynamics across multiple temporal resolutions, yet effectively incorporating multi-scale information into generative modeling remains challenging without increasing generative complexity. We propose \method{}, a unified framework that integrates multi-scale representation learning with efficient non-autoregressive generation through a shared-semantic discrete interface. Specifically, a multi-scale tokenizer employs shared-index voting to consolidate complementary temporal information into a single public-ID sequence, while scale-specific codebooks preserve the information required for reconstruction. Building on this unified representation, we develop geometry-aware categorical flow matching, which initializes transport from learned low-rank anchors and generates public identities non-autoregressively. This design captures multi-scale temporal structure without requiring multiple generative token streams or sequential token decoding. Experiments across four benchmarks and both short- and long-term settings demonstrate improved generative fidelity, strong cross-scale consistency, and substantial inference speedups over representative autoregressive and diffusion-based baselines. Code is available at \url{https://github.com/kkking-kk/UniFlow}
\end{abstract}


    \section{Introduction}
    \label{sec:intro}
    
    Time-series generation aims to learn the underlying
    distribution of temporal processes, supporting applications
    such as simulation, data augmentation, privacy-preserving
    analysis, and downstream model development. Recent advances
    span adversarial generators~\citep{yoon2019time,xu2020cot},
    variational latent models~\citep{desai2021timevae},
    diffusion-based approaches~\citep{tashiro2021csdi,alcaraz2022diffusion,yuan2024diffusion},
    and flow-based generative models~\citep{hu2025flowts}.
    Despite their progress in generative fidelity, effectively
    incorporating multiscale temporal information into a unified
    and efficient generation process remains an important challenge.
    
    Real-world time series exhibit complex dynamics across
    multiple temporal resolutions, ranging from local
    fluctuations and periodic variations to long-term trends.
    These structures provide complementary information about
    the same underlying process, motivating multiscale
    representation learning in temporal modeling
    ~\citep{lim2021time,wu2020adversarial,ding2020hierarchical}.
    For generation, however, capturing individual temporal
    patterns is not sufficient: the generated sequences
    must also preserve their relationships across scales.
    This raises a fundamental question:
    \emph{How can complementary multiscale information be
    integrated into a unified generative representation
    without introducing multiple token streams or excessive
    generation complexity?}
    Addressing this question involves three key challenges.
    
    \noindent\textbf{1. Capturing Complementary Multi-scale Dynamics.}
    Temporal patterns exhibit different characteristics
    across resolutions, making it difficult for a single
    temporal view to explicitly distinguish local variations
    from broader temporal structures. An effective
    representation should capture complementary information
    at different scales while preserving their correspondence
    to the same underlying sequence.
    
    \noindent\textbf{2. Unifying Multi-scale Semantics without Interface Expansion.}
    Multi-scale representations naturally introduce multiple
    sets of latent features or discrete tokens. Directly
    generating separate scale-specific streams requires
    coordination across representations, while concatenating
    them increases the sequence length presented to the
    generator. The challenge is therefore to consolidate
    multi-scale information into shared semantic identities
    without expanding the generator-facing token sequence.
    
    \noindent\textbf{3. Efficient Generation of Unified Token Sequences.}
    Autoregressive token generation relies on sequential
    decoding, where predictions depend on previously
    generated tokens. Its teacher-forced training procedure
    also introduces a discrepancy between training and
    inference histories, potentially accumulating errors
    during generation~\citep{bengio2015scheduled,schmidt2019generalization}.
    An alternative should generate the complete shared-token
    sequence non-autoregressively while effectively modeling
    the structure of its latent distribution.
    
    To address these challenges, we propose \method{},
    a unified multi-scale time-series generation framework
    that connects shared-semantic tokenization with
    geometry-aware categorical flow matching.
    In Stage 1, aligned scale-specific encoders extract
    complementary temporal representations, while
    shared-index voting assigns a common public identity
    to each temporal position. Scale-specific codebooks
    retain reconstruction details, whereas the resulting
    public-ID sequence provides a single, length-$L$
    generative interface regardless of the number of scales.
    In Stage 2, a learned low-rank anchor prior provides
    structured initialization for categorical flow matching,
    which models the complete public-ID trajectory through
    non-autoregressive continuous transport. This design
    integrates multi-scale representation learning with
    efficient generation without requiring multiple
    synchronized token streams.
    
    Our main contributions are summarized as follows:
    \begin{itemize}
        \item \textbf{Unified Multi-scale Semantic Tokenization.}
        We introduce shared-index voting with scale-specific
        codebooks to integrate complementary temporal
        information into a single public-ID sequence,
        preserving multi-scale reconstruction information
        without expanding the generative interface.
    
        \item \textbf{Geometry-Aware Non-Autoregressive Generation.}
        We develop a categorical flow framework that
        combines learned low-rank anchor initialization
        with parallel public-ID generation, avoiding
        sequential token decoding while leveraging
        the learned latent structure.
    
        \item \textbf{Generative Fidelity and Multi-scale Consistency.}
        Extensive experiments across four benchmarks
        demonstrate improved generative fidelity and
        inference efficiency. Reconstruction, cross-scale
        consistency, and scale-aware statistical analyses
        further validate the effectiveness of the
        unified multi-scale representation.
    \end{itemize}

\section{Related Work}
	\label{sec:related}
	
    \noindent\textbf{Generative Models for Time Series Generation.}
    Generative modeling has advanced time series generation
    through diverse paradigms, including adversarial learning,
    variational inference, diffusion, and flow-based methods.
    TimeGAN~\citep{yoon2019time} and COT-GAN~\citep{xu2020cot}
    employ adversarial objectives to capture temporal
    distributions, while TimeVAE~\citep{desai2021timevae}
    provides a latent-variable generative framework.
    Diffusion-based methods~\citep{tashiro2021csdi, alcaraz2022diffusion,yuan2024diffusion,feng2024latent}
    achieve expressive distribution modeling through
    iterative denoising, whereas FlowTS~\citep{hu2025flowts}
    explores efficient continuous transport via rectified flow.
    Despite their respective strengths, effectively
    incorporating intrinsic temporal characteristics,
    particularly multiscale dependencies, into generative
    modeling while balancing generative fidelity and
    sampling efficiency remains an important research direction.
	

    \noindent\textbf{Discrete and Multi-scale Representation Learning.}
    VQ-VAE~\citep{van2017neural} and VQGAN~\citep{esser2021taming}
    compress observations into reusable discrete tokens.
    Hierarchical and residual quantization further enhance
    representational capacity through multiple latent levels~\citep{williams2020hierarchical, feng2025hdt}.
    Recently, vector-quantized modeling has gained increasing
    attention in time series generation. SDformer~\citep{chen2024sdformer} and SDFlow~\citep{li2026sdflow}
    introduces similarity-driven discrete token modeling,
    while MSDformer~\citep{feng2026msdformer} and
    TimeMAR~\citep{xu2026timemar} explore multi-scale discrete
    generation.
    Despite these advances, balancing temporal representation
    quality and efficient generative fidelity remains
    an important challenge. \method{} addresses this challenge
    by integrating multi-scale information into a unified
    public-token interface for non-autoregressive generation.

    \noindent\textbf{Flow Matching and Non-Autoregressive Generation.}
    Flow matching and rectified flow learn continuous transport fields~\citep{lipman2023flow,liu2023flow,
    tong2024improving}, while stochastic interpolants connect
    flow-based and diffusion-based modeling~\citep{albergo2023stochastic,chen2024flow}.
    Variational flow matching further supports flexible
    endpoint distributions, including categorical posteriors~\citep{eijkelboom2024variational}, with DiT providing
    a scalable generative backbone~\citep{peebles2023scalable}.
    In discrete representation learning, TokenFlow~\citep{qu2024tokenflow} aligns text-guided semantic
    and pixel-level visual features through shared code indices,
    unifying image understanding and generation.
    In contrast, \method{} employs shared-index learning
    to integrate multi-scale temporal evidence into a single
    public-ID sequence, enabling non-autoregressive categorical
    flow generation without separate scale-specific token streams.

\noindent\textbf{The Position of UniFlow}
Existing studies have advanced time series generation
through expressive generative models, multiscale discrete
representations, and efficient flow-based learning.
However, effectively integrating complementary temporal
information without increasing generative complexity
remains an important challenge. To bridge these directions,
\method{} unifies multiscale semantic representations
through a shared public-token interface and employs
geometry-aware categorical flow matching for
non-autoregressive generation. This design enables
rich multiscale modeling within a single generative
sequence, providing an effective balance between
representation capacity and efficient generative fidelity.


    \section{Preliminaries}
	\label{sec:prelim}
	
	We fix the notation shared by both stages and briefly recall the single-scale components that \method{} generalizes. UniFlow combines a shared-index multi-scale token interface with non-autoregressive token space generation.
	
	\noindent\textbf{Problem Formulation.}
	Let $\mat{X}_{1:\ell}=(\vect{x}_1,\ldots,\vect{x}_\ell)\in\R^{\ell\times d}$ be a multivariate time series drawn from the unknown data distribution $p_{\mathrm{data}}$. Given a source variable $\vect{Z}_0\sim\pi_0$, unconditional generation learns a mapping $\mathcal{G}$ such that
	\begin{equation}
		\hat{\mat{X}}_{1:\ell}=\mathcal{G}(\vect{Z}_0),
		\qquad
		p_{\mathcal{G}}\approx p_{\mathrm{data}}.
	\end{equation}
	For aligned temporal views $\{\mathcal{T}_m\}_{m=1}^{M}$ induced by different receptive fields or patch scales, the multi-scale objective additionally requires the joint generated view distribution
	\begin{equation}
		\operatorname{Law}\!\left(
		\mathcal{T}_1(\hat{\mat{X}}),\ldots,
		\mathcal{T}_M(\hat{\mat{X}})
		\right)
	\end{equation}
	to approximate the corresponding joint distribution of real views. The goal is therefore one coherent generated sequence, rather than independently generated outputs for different temporal scales.
	
	\noindent\textbf{Similarity-Based Quantization.}
	Following similarity-based discrete tokenization~\citep{chen2024sdformer, li2026sdflow}, a VQ tokenizer first maps the input sequence to latent features $\mat{H}_{1:L} = \mathcal{E}(\mat{X}_{1:\ell}) \in \R^{L \times d_c}$, where $L=\ell/s$ is the compressed token length. Each latent vector is then assigned to one code in a learned vocabulary $\mathcal{C} = \{\vect{c}_k\}_{k=0}^{K-1} \subset \R^{d_c}$ by cosine similarity:
	\begin{equation}
		\label{eq:similarity_driven}
		y_i = Q(\vect{h}_i) := \arg\max_{k=0,\ldots,K-1} \frac{\vect{h}_i}{\|\vect{h}_i\|} \cdot \frac{\vect{c}_k}{\|\vect{c}_k\|}.
	\end{equation}
	Normalized similarities help prevent the codebook from collapsing toward a few large-norm entries, after which the decoder reconstructs the series from the retrieved embeddings. \method{} extends this rule with aligned branches that produce scale-specific latents but jointly export one public token sequence and codebook; Section~\ref{sec:vqvae} details the shared mapping across branch-specific codebooks.
	
	\noindent\textbf{Rectified Flow Matching (RFM)~\citep{liu2022flow}.} RFM learns a time-dependent velocity field $\vect{v}_\theta: \R^D \times [0,1] \to \R^D$ whose ODE moves samples from a simple source distribution toward the target distribution:
	\begin{equation}
		\frac{d\vect{z}_t}{dt} = \vect{v}_\theta(\vect{z}_t, t), \quad \vect{z}_0 \sim p_0, \quad \vect{z}_1 \sim p_{\text{data}}.
	\end{equation}
	For the linear interpolation path $\vect{z}_t = (1-t)\vect{z}_0 + t\vect{z}_1$, the target velocity becomes $\vect{z}_1 - \vect{z}_0$, leading to the objective
	\begin{equation}
		\label{eq:continuous_fm}
		\mathcal{L}_{\text{FM}} = \E_{t, \vect{z}_0, \vect{z}_1}\left[ \|\vect{v}_\theta(\vect{z}_t, t) - (\vect{z}_1 - \vect{z}_0)\|^2 \right].
	\end{equation}
	Because the whole sequence evolves jointly under this ODE, flow matching removes the left-to-right dependency that drives exposure bias in autoregressive token models.

\section{Method}
\label{sec:method}

\begin{figure*}[t]
    \centering
    \includegraphics[width=0.95\textwidth]{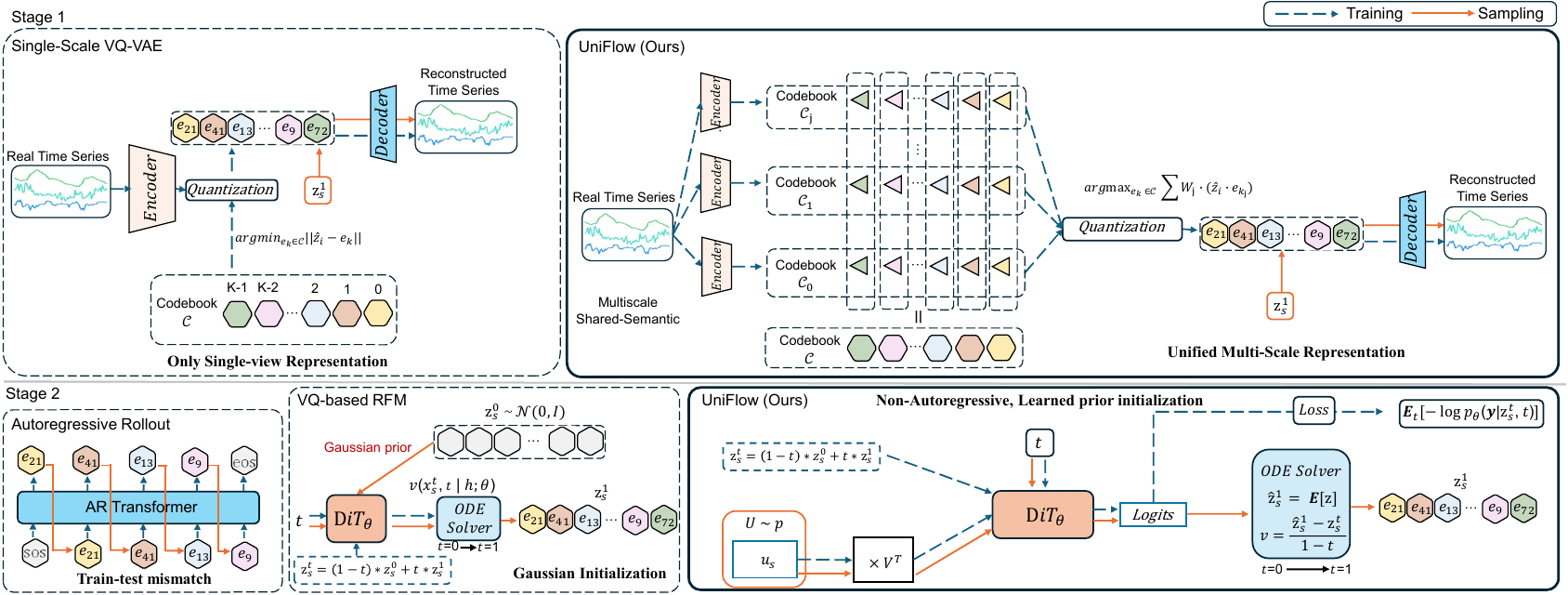}
    \caption{Overview of \method{}. Stage~1 integrates aligned temporal representations through shared-index voting, yielding a single public-ID sequence and scale-specific codebooks. Stage~2 generates public identities using anchor-initialized categorical flow matching, followed by scale-specific embedding retrieval and decoding. Dashed blue and solid orange arrows denote training and sampling, respectively.}
    \label{fig:architecture}
\end{figure*}

As illustrated in Fig.~\ref{fig:architecture}, \method{} consists of two stages that connect multiscale representation learning with unified non-autoregressive generation. Stage~1 learns a shared-semantic tokenizer, in which aligned temporal views jointly determine a single public-ID sequence while retaining scale-specific codebooks for reconstruction. Stage~2 models this sequence through geometry-aware categorical flow matching, combining a learned low-rank anchor prior with non-autoregressive endpoint prediction. Thus, multiscale information is incorporated into a single generative interface without introducing separate token streams for individual scales.

\subsection{Stage 1: Shared-Semantic Multiscale Tokenization}
\label{sec:vqvae}

\noindent\textbf{Aligned Temporal Representations.} Given an input time series $\mat{X}_{1:\ell}\in\R^{\ell\times d}$, we employ $M$ temporal encoders with different receptive fields or patch sizes:
\begin{equation}
    \mat{H}^{(m)}_{1:L}
    =\mathcal{E}^{(m)}(\mat{X}_{1:\ell})
    \in\R^{L\times d_m},
    \qquad \sum_{m=1}^{M}d_m=d_c.
\end{equation}
All branches are aligned to the same compressed temporal length $L$, so each position $i$ is characterized by the complementary representations $\{\vect{h}^{(m)}_i\}_{m=1}^{M}$. Our implementation uses fine- and coarse-scale branches, while the formulation allows more temporal resolutions.

Independently quantizing these views would produce multiple scale-specific token sequences, requiring the generator to represent their temporal evolution and cross-scale dependencies. We instead separate \emph{shared semantic identity} from \emph{scale-specific realization}: the branches retain private embedding spaces but jointly select a public discrete index.

\noindent\textbf{(1) Shared-index Voting Quantization.}
Branch $m$ maintains a codebook
\begin{equation}
    \mathcal{C}^{(m)}
    =\{\vect{c}^{(m)}_k\}_{k=0}^{K-1}
    \subset\R^{d_m},
\end{equation}
with a common index set $\{0,\ldots,K-1\}$ across branches. For each position $i$ and candidate code $k$, the normalized branch-wise similarity is
\begin{equation}
    s_{i,k}^{(m)}
    =
    \frac{\vect{h}^{(m)}_i}{\|\vect{h}^{(m)}_i\|_2}
    \cdot
    \frac{\vect{c}^{(m)}_k}{\|\vect{c}^{(m)}_k\|_2}.
\end{equation}
The branch-wise evidence is combined before discrete assignment:
\begin{equation}
\label{eq:shared_mapping}
y_i = \arg\max_{k=0,\ldots,K-1} \sum_{m=1}^{M} \alpha_m s_{i,k}^{(m)},
	\qquad \sum_{m=1}^{M} \alpha_m = 1.
\end{equation}
Accordingly, each temporal position receives one public ID rather than separate scale-specific assignments. Importantly, a shared index does not force the branch embeddings to be identical: code $k$ acts as a common semantic key whose realization $\vect{c}^{(m)}_k$ remains specific to scale $m$.

\noindent\textbf{(2) Public-token Construction and Training.}
After selecting $y_i$, each branch retrieves its private codeword $\tilde{\vect{h}}^{(m)}_i=\vect{c}^{(m)}_{y_i}$. The decoder reconstructs the sequence from the concatenated embeddings
\begin{equation}
    \tilde{\vect{h}}_i
    =\operatorname{concat}
    \big(\tilde{\vect{h}}^{(1)}_i,\ldots,
    \tilde{\vect{h}}^{(M)}_i\big)
    \in\R^{d_c}.
\end{equation}
This induces a fused public codebook
\begin{equation}
    \label{eq:shared_codebook}
    \mathcal{C}_{\mathrm{shared}}
    =\left\{
    \vect{c}^{\mathrm{shared}}_k
    =\operatorname{concat}\big(
    \vect{c}^{(1)}_k,\ldots,
    \vect{c}^{(M)}_k\big)
    \right\}_{k=0}^{K-1}.
\end{equation}
Stage~1 is trained with reconstruction and branch-wise commitment losses:
\begin{equation}
\begin{aligned}
    \mathcal{L}_{\mathrm{VQ}}
    &=\|\mat{X}_{1:\ell}-\tilde{\mat{X}}_{1:\ell}\|_F^2
    \\
    &\quad+\frac{\lambda}{L}
    \sum_{i=1}^{L}\sum_{m=1}^{M}\alpha_m
    \left(1-
    \vect{h}^{(m)}_i\cdot
    \operatorname{sg}(\tilde{\vect{h}}^{(m)}_i)
    \right),
\end{aligned}
\end{equation}
where the dot products are evaluated using normalized branch features and codewords, and $\operatorname{sg}(\cdot)$ denotes stop-gradient. Scale-specific codebooks are updated by exponential moving average (EMA), with inactive-code resetting to maintain utilization.

Once trained, the tokenizer is frozen. Its output consists of one public-ID sequence $y_{1:L}$ and the fused endpoint representation
\begin{equation}
    \label{eq:public_token_target}
    \mat{Z}_1
    =\big[
    \vect{c}^{\mathrm{shared}}_{y_1},\ldots,
    \vect{c}^{\mathrm{shared}}_{y_L}
    \big]^\top\in\R^{L\times d_c},
\end{equation}
which provides the shared discrete vocabulary and continuous endpoint targets for Stage~2.

\subsection{Stage 2: Geometry-Aware Categorical Flow Matching}
\label{sec:vq-vfm}

Stage~2 models the public-token sequence exported by the frozen tokenizer. We first construct a structured source distribution informed by the geometry of training endpoints, and then learn a continuous velocity field through categorical prediction over their public identities.

\noindent\textbf{(1) Low-rank Anchor Initialization.}
Valid public-token endpoints are assembled from a finite shared codebook and exhibit sequence-level structure in the ambient latent space. To capture this geometry, we vectorize the endpoints as
\begin{equation}
    \vect{z}_1=\operatorname{vec}(\mat{Z}_1)\in\R^D,
    \qquad D=Ld_c,
\end{equation}
and collect $N_a$ training endpoints in
\begin{equation}
    \mat{Z}_a
    =\big[\vect{z}_1^{(1)},\ldots,
    \vect{z}_1^{(N_a)}\big]^\top
    \in\R^{N_a\times D}.
\end{equation}
We approximate these endpoints using a learned low-rank factorization,
\begin{equation}
    \mat{Z}_a\approx\mat{U}\mat{V}^\top,
    \qquad r\ll D,
\end{equation}
where $\mat{U}\in\R^{N_a\times r}$ contains support coordinates and $\mat{V}\in\R^{D\times r}$ provides a shared basis. The coordinates are regularized by
\begin{equation}
    \label{eq:anchor_regularization}
    \mathcal{L}_{\mathrm{reg}}
    =\lambda_\mu\|\bar{\vect{u}}\|_2^2
    +\lambda_\sigma
    \left|\operatorname{std}(\mat{U})-1\right|.
\end{equation}
For a sampled support coordinate $\vect{u}_a$, we construct an anchor by local perturbation:
\begin{equation}
    \vect{\epsilon}\sim\mathcal{N}(\vect{0},h^2\mat{I}_r),
    \qquad
    \vect{z}_0=(\vect{u}_a+\vect{\epsilon})\mat{V}^\top.
\end{equation}
After reshaping $\vect{z}_0$ into $\mat{Z}_0\in\R^{L\times d_c}$, we apply token-wise normalization,
\begin{equation}
    \mat{Z}_0\leftarrow\operatorname{RowNorm}(\mat{Z}_0).
\end{equation}
The bandwidth $h$ is determined from neighboring support-coordinate distances. This procedure defines an anchor distribution $p_{\mathrm{anchor}}$ concentrated around the learned low-rank scaffold rather than an isotropic ambient-space source.

\noindent\textbf{(2) Categorical Endpoint Prediction.}
For an anchor--endpoint training pair $(\mat{Z}_0,\mat{Z}_1)$ and $t\in[0,1)$, the interpolation is
\begin{equation}
    \label{eq:uniflow_interpolation}
    \mat{Z}_t=(1-t)\mat{Z}_0+t\mat{Z}_1.
\end{equation}
A DiT-based network predicts logits $f_\theta(\mat{Z}_t,t)\in\R^{L\times K}$, defining a categorical posterior at each position:
\begin{equation}
\begin{aligned}
    q_\theta(y_i=k\mid\mat{Z}_t,t)
    &=\pi_{\theta,i}^{t,k},\\
    \pi_{\theta,i}^{t,k}
    &=\frac{\exp(f_{\theta,i}^{k}(\mat{Z}_t,t)/\tau)}
    {\sum_{j=0}^{K-1}\exp(f_{\theta,i}^{j}(\mat{Z}_t,t)/\tau)},
\end{aligned}
\end{equation}
where $\tau$ is the softmax temperature. Rather than regressing an unconstrained endpoint embedding, the model directly learns the probabilities of the valid public identities supplied by Stage~1.

\noindent\textbf{(3) Categorical-to-continuous Transport.}
The categorical prediction is mapped to the shared embedding space using its expected codeword:
\begin{equation}
    \label{eq:uniflow_endpoint_mean}
    \boldsymbol{\mu}_{\theta,i}(\mat{Z}_t,t)
    =\sum_{k=0}^{K-1}\pi_{\theta,i}^{t,k}
    \vect{c}^{\mathrm{shared}}_k.
\end{equation}
Stacking these means gives $\mat{\mathrm{M}}_\theta(\mat{Z}_t,t)\in\R^{L\times d_c}$. We parameterize the velocity field by
\begin{equation}
    \label{eq:uniflow_velocity}
    \mat{V}_\theta(\mat{Z}_t,t)
    =\frac{\mat{\mathrm{M}}_\theta(\mat{Z}_t,t)-\mat{Z}_t}{1-t},
    \qquad t<1.
\end{equation}
The categorical cross-entropy objective is
\begin{equation}
    \label{eq:categorical_ce_main}
    \mathcal{L}_{\mathrm{CE}}
    =-\mathbb{E}_{t,(\mat{Z}_0,\mat{Z}_1)}
    \left[\sum_{i=1}^{L}
    \log\pi_{\theta,i}^{t,y_i}\right],
\end{equation}
where the expectation is taken under the anchor--endpoint training coupling. The categorical objective trains the flow network, while $\mathcal{L}_{\mathrm{reg}}$ regularizes the learned anchor coordinates. The two roles are distinct: the anchor prior specifies the source geometry, whereas the categorical network predicts endpoint identities and thereby determines continuous transport.

\noindent\textbf{(4) Non-autoregressive Sampling and Decoding.}
At inference, we sample $\mat{Z}_0\sim p_{\mathrm{anchor}}$ and numerically solve
\begin{equation}
    \frac{\mathrm{d}\mat{Z}_t}{\mathrm{d}t}
    =\mat{V}_\theta(\mat{Z}_t,t)
\end{equation}
to obtain an estimated terminal state $\widehat{\mat{Z}}_1$. Public identities are retrieved position-wise using cosine similarity:
\begin{equation}
    \label{eq:uniflow_inference_readout}
    \hat y_i
    =\arg\max_{k\in\{0,\ldots,K-1\}}
    \cos\left(\hat{\vect{z}}_{1,i},
    \vect{c}^{\mathrm{shared}}_k\right).
\end{equation}
Each generated ID retrieves its private codewords $\{\vect{c}^{(m)}_{\hat y_i}\}_{m=1}^{M}$, which are concatenated and passed to the frozen decoder to produce $\widehat{\mat{X}}_{1:\ell}$. The ODE evolves all temporal positions jointly, avoiding sequential autoregressive token decoding while retaining a generator-facing sequence of length $L$.


\noindent\textbf{Overall Training and Generation Workflow.}
\method{} follows a two-stage training procedure.
Stage 1 learns the shared-semantic tokenizer and
scale-specific codebooks, which are subsequently frozen
to provide unified public token representations for
Stage 2. The categorical flow generator then learns
to transport samples from the structured anchor prior
toward these representations. During inference,
the generated latent states are mapped to public IDs,
which retrieve scale-specific embeddings for
time series reconstruction. This workflow integrates
multi-scale representation learning with non-autoregressive
generation through a single public-token interface.

\subsection{Theoretical Properties of Unified Multiscale Generation}
\label{sec:theory}

We analyze the unified interface from two perspectives. At the representation level of \textbf{Stage 1}, we examine the reliability of shared voting quantization, the recoverability of public identities from individual views, and the complexity of length-preserving generation. At the generative level of \textbf{Stage 2}, we characterize the geometry of the anchor source and the consistency between categorical endpoint prediction and continuous velocity learning.

\paragraph{Reliability of shared-index voting.}
 The shared mapping raises a fundamental question: \textbf{when does aggregating multiple temporal views yield a more reliable semantic assignment?} Let $y_i^\star$ be a fixed reference identity, specified independently of the score fluctuations being analyzed. For a competing code $k\neq y_i^\star$, define the branch-wise target-versus-competitor margin
\begin{equation}
    \Delta_{i,k}^{(m)}
    =s_{i,y_i^\star}^{(m)}-s_{i,k}^{(m)}.
\end{equation}
Writing the vector of margins as
\begin{equation}
    \boldsymbol{\Delta}_{i,k}
    =\boldsymbol{\mu}_{i,k}+\boldsymbol{\varepsilon}_{i,k},
\end{equation}
we regard $\boldsymbol{\mu}_{i,k}$ as the expected cross-scale margin and $\boldsymbol{\varepsilon}_{i,k}$ as its variation.

\begin{proposition}[Robustness of shared-index voting]
\label{prop:shared_vote_robustness}
Suppose $\boldsymbol{\varepsilon}_{i,k}$ is zero-mean sub-Gaussian with covariance proxy $\boldsymbol{\Sigma}_{i,k}\succeq0$ and $\boldsymbol{\alpha}^\top\boldsymbol{\mu}_{i,k}>0$ for all $k\neq y_i^\star$. Then the shared assignment $\widetilde y_i$ satisfies
\begin{equation}
    \label{eq:shared_vote_error_bound_main}
    \mathbb{P}(\widetilde y_i\neq y_i^\star)
    \leq\sum_{k\neq y_i^\star}
    \exp\left(
    -\frac{(\boldsymbol{\alpha}^\top\boldsymbol{\mu}_{i,k})^2}
    {2\boldsymbol{\alpha}^\top\boldsymbol{\Sigma}_{i,k}
    \boldsymbol{\alpha}}
    \right),
\end{equation}
where a term with zero variance proxy and a positive projected margin is interpreted as zero.
\end{proposition}

The result follows from a sub-Gaussian tail bound on each aggregated margin and a union bound over competing codes (Appendix~\ref{app:shared_margin}). It shows that shared-vote reliability is governed by the aggregated margin relative to cross-scale uncertainty. Additional temporal views can therefore improve assignment reliability when their complementary evidence increases this margin-to-uncertainty ratio. This result characterizes the conditions under which
aggregating cross-scale evidence can improve assignment
reliability, providing a probabilistic motivation
for shared-index voting.


\paragraph{Single-view recoverability and decoder stability.}
We next examine whether the public ID $y_i$ from Eq.~\eqref{eq:shared_mapping} remains recognizable from an individual view. For branch $r$, define
\begin{equation}
    y_{i,\mathrm{rec}}^{(r)}=\arg\max_k s_{i,k}^{(r)},
    \qquad
    \gamma_i^{(r)}
    =s_{i,y_i}^{(r)}-\max_{k\neq y_i}s_{i,k}^{(r)}.
\end{equation}

\begin{proposition}[Single-view recovery and decoder stability]
\label{prop:single_view_recovery_main}
If $\gamma_i^{(r)}>0$, branch $r$ uniquely recovers the shared identity, i.e., $y_{i,\mathrm{rec}}^{(r)}=y_i$. This assignment remains unchanged under similarity-score perturbations bounded by $\epsilon$ in $\ell_\infty$ norm whenever $\gamma_i^{(r)}>2\epsilon$.

Further, suppose the decoder $\mathcal{D}_{\mathrm{dec}}$ is $L_{\mathrm{dec}}$-Lipschitz, and define the diameter of branch codebook $m$ by
\begin{equation}
    B_m=\max_{j,k}\|\vect{c}^{(m)}_j-\vect{c}^{(m)}_k\|_2.
\end{equation}
Let $\mat{H}_{\mathrm{shared}}$ and $\mat{H}_{\mathrm{rec}}$ denote the fused codeword matrices selected by the shared and recovered IDs, respectively, and let $S$ be the set of positions where these IDs differ. Then
\begin{equation}
    \label{eq:decoder_stability_main}
\begin{aligned}
    &\|\mathcal{D}_{\mathrm{dec}}(\mat{H}_{\mathrm{rec}})
    -\mathcal{D}_{\mathrm{dec}}(\mat{H}_{\mathrm{shared}})\|_F
    \\
    &\quad\leq L_{\mathrm{dec}}
    \sqrt{|S|\sum_{m=1}^{M}B_m^2}.
\end{aligned}
\end{equation}
Consequently, for an input sequence $\mat{X}$,
\begin{equation}
    \label{eq:reconstruction_error_bound}
\begin{aligned}
    &\|\mat{X}-\mathcal{D}_{\mathrm{dec}}(\mat{H}_{\mathrm{rec}})\|_F
    \\
    &\quad\leq
    \|\mat{X}-\mathcal{D}_{\mathrm{dec}}(\mat{H}_{\mathrm{shared}})\|_F
    +L_{\mathrm{dec}}\sqrt{|S|\sum_{m=1}^{M}B_m^2}.
\end{aligned}
\end{equation}
\end{proposition}

The margin conditions ensure the stability of individual code assignments; the decoder bounds follow from codebook diameters, decoder Lipschitzness, and the triangle inequality (Appendix~\ref{app:shared_margin}). They clarify how a common public identity can be recovered from scale-specific representations, and how mismatched identities can affect reconstruction.

\noindent\textbf{Empirical Verification of Shared-ID Recoverability.} Table~\ref{tab:main_shared_id_match} complements the above analysis by evaluating how well the canonical public-ID sequence obtained from the full shared mapping can be recovered from an individual temporal view. The Shared Ref. columns report agreement with the original shared assignment itself and therefore equal $1.0$ by construction, serving as the reference for the complete multi-scale representation. In contrast, using only Scale 1 or Scale 2 recovers between $78.73\%$ and $90.53\%$ of the public IDs across the four datasets. This indicates that the shared semantic identity remains largely recognizable from either temporal resolution.
    
Importantly, the single-view agreement is high but not perfect, suggesting that the two scales preserve complementary rather than fully redundant information. The lower sequence-level agreement is expected because it requires exact recovery of every token in a sequence; consequently, even a small number of position-wise mismatches causes the entire sequence to be counted as inconsistent.

    
    \begin{table}[t]
    	\centering
    	\small
    	\resizebox{1.0\columnwidth}{!}{
    		\begin{tabular}{@{}lcccccc@{}}
    			\toprule
    			\textbf{Dataset} 
    			& \textbf{Shared Ref. Token} 
    			& \textbf{Shared Ref. Seq} 
    			& \textbf{Scale-1 Token} 
    			& \textbf{Scale-1 Seq} 
    			& \textbf{Scale-2 Token} 
    			& \textbf{Scale-2 Seq} \\
    			\midrule
    			Sine   & 1.0000 & 1.0000 & 0.7873 & 0.2396 & 0.8540 & 0.3906 \\
    			Stock  & 1.0000 & 1.0000 & 0.8053 & 0.3372 & 0.9053 & 0.5789 \\
    			ETTh   & 1.0000 & 1.0000 & 0.8305 & 0.3389 & 0.8676 & 0.4281 \\
    			Energy & 1.0000 & 1.0000 & 0.8847 & 0.4974 & 0.8823 & 0.4868 \\
    			\bottomrule
    		\end{tabular}
    	}
    	\caption{Agreement with the canonical public-ID sequence. Shared Ref. denotes the full shared mapping, while Scale 1 and Scale 2 recover the public IDs from a single temporal view.}
    	\label{tab:main_shared_id_match}
    \end{table}

\paragraph{Complexity of the unified interface.}
The public representation exposes $L$ tokens to Stage~2 independently of the number of aligned temporal branches.

\begin{proposition}[Complexity of the unified interface]
\label{prop:interface_complexity_main}
Assume $M$ aligned branches, each with $L$ tokens, and a fixed per-token DiT hidden width $d_{\mathrm{model}}$. The unified public-token sequence requires a self-attention term of
\begin{equation}
    O(L^2d_{\mathrm{model}}).
\end{equation}
Concatenating $M$ separate scale-specific sequences yields $ML$ tokens and a self-attention term of
\begin{equation}
    O(M^2L^2d_{\mathrm{model}}),
\end{equation}
while $M$ separate attention streams require a total of
\begin{equation}
    O(ML^2d_{\mathrm{model}}),
\end{equation}
before any cross-stream interaction.
\end{proposition}

Its structural implication is that a shared public-ID interface avoids the scale-dependent expansion of the generator-facing token sequence.

\paragraph{Structured Initialization and Categorical-flow Consistency.}
The learned anchor prior provides a structured source
distribution for categorical flow transport. Specifically,
for the unnormalized anchor
$\vect{z}_0^{\mathrm{pre}}
=(\vect{u}_a+\vect{\epsilon})\mat{V}^\top$,
where $\vect{\epsilon}\sim\mathcal{N}(\vect{0},h^2\mat{I}_r)$,
we have
\begin{equation}
\begin{aligned}
    \mathbb{E}[\vect{z}_0^{\mathrm{pre}}\mid\vect{u}_a]
    &=\vect{u}_a\mat{V}^\top,\\
    \operatorname{Cov}(\vect{z}_0^{\mathrm{pre}}\mid\vect{u}_a)
    &=h^2\mat{V}\mat{V}^\top.
\end{aligned}
\end{equation}
Thus, the source perturbations before row normalization
lie within a learned subspace of dimension at most $r$.
Given the resulting anchor--endpoint coupling,
the population conditional velocity associated with
Eq.~\eqref{eq:uniflow_interpolation} is
\begin{equation}
    \mat{V}^{\star}(\mat{Z}_t,t)
    =
    \mathbb{E}[
    \mat{Z}_1-\mat{Z}_0\mid\mat{Z}_t,t
    ].
\end{equation}
We next establish how categorical endpoint learning
recovers this conditional velocity field under
the population-optimality assumption.



\begin{proposition}[Categorical endpoint and velocity consistency]
\label{prop:categorical_consistency_main}
Assume the categorical model can represent the true conditional position-wise distributions and that the population cross-entropy optimum is attained. Then, for each temporal position $i$,
\begin{equation}
    q_\theta(y_i=k\mid\mat{Z}_t,t)
    =p(y_i=k\mid\mat{Z}_t,t).
\end{equation}
Consequently,
\begin{equation}
    \boldsymbol{\mu}_{\theta,i}(\mat{Z}_t,t)
    =\mathbb{E}[\vect{z}_{1,i}\mid\mat{Z}_t,t],
\end{equation}
and the induced velocity field satisfies, for $t<1$,
\begin{equation}
    \mat{V}_\theta(\mat{Z}_t,t)
    =\mat{V}^{\star}(\mat{Z}_t,t).
\end{equation}
\end{proposition}

At the population optimum, the position-wise categorical probabilities recover the conditional codebook expectations. Moreover, interpolation implies
\begin{equation}
\begin{aligned}
    &\mathbb{E}[\mat{Z}_1-\mat{Z}_0\mid\mat{Z}_t,t]
    \\
    &\quad=
    \frac{\mathbb{E}[\mat{Z}_1\mid\mat{Z}_t,t]-\mat{Z}_t}{1-t},
\end{aligned}
\end{equation}
which coincides with Eq.~\eqref{eq:uniflow_velocity}. Thus, categorical endpoint supervision is consistent with the corresponding conditional continuous velocity field under the stated assumptions. The conclusion concerns position-wise conditional marginals and does not, on its own, establish consistency of the full joint categorical sequence distribution.



\paragraph{Consistency of geometric readout.}
The final public-ID assignment is obtained through
cosine similarity between the generated terminal
state and the shared codebook. For nonzero vectors,
let $\bar{\vect{z}}_{1,i}$ and $\bar{\vect{c}}_k$
denote their normalized forms, and define
\begin{equation}
    \delta_k^{\mathrm{cos}}
    =
    \min_{j\neq k}
    \|\bar{\vect{c}}_k-\bar{\vect{c}}_j\|_2.
\end{equation}
Assuming distinct normalized codewords, if
\begin{equation}
    \|\bar{\vect{z}}_{1,i}
    -\bar{\vect{c}}_k\|_2
    <
    \frac{\delta_k^{\mathrm{cos}}}{2},
\end{equation}
then $k$ is the unique cosine nearest-code
assignment. This follows from the triangle
inequality and the equivalence between cosine
maximization and Euclidean nearest-neighbor
search on normalized vectors.

This condition provides a sufficient guarantee for stable public-ID retrieval from the generated terminal state, establishing a geometric foundation for reliable discrete token reconstruction in \method{}.

\begin{algorithm}[t]
    \caption{Training procedure of \method{}.}
    \label{alg:app_pipeline}
    \begin{algorithmic}[1]
        \REQUIRE Training set $\mathcal{D}$, multi-scale encoders,
        branch codebooks, decoder, and voting weights.
        \STATE \textbf{Stage 1: shared-semantic tokenization}
        \FOR{each minibatch $\mat{X}\sim\mathcal{D}$}
            \STATE Encode aligned temporal views and compute branch-wise similarities.
            \STATE Select each public ID by the weighted shared-mapping rule.
            \STATE Retrieve private branch embeddings and concatenate them.
            \STATE Reconstruct $\mat{X}$ and optimize $\mathcal{L}_{\mathrm{VQ}}$.
            \STATE Update branch codebooks by EMA and inactive-code resetting.
        \ENDFOR
        \STATE Freeze Stage 1 and construct public-token matrices $\mat{Z}_1$.
        \STATE Vectorize public-token matrices and initialize
        $\mat{U},\mat{V}$ from $\mat{Z}_a\approx\mat{U}\mat{V}^{\top}$.
        \STATE \textbf{Stage 2: anchor-initialized categorical flow}
        \FOR{each endpoint minibatch}
            \STATE Construct $\mat{Z}_0$ from the corresponding low-rank coordinates.
            \STATE Sample $t$ and set
            $\mat{Z}_t=(1-t)\mat{Z}_0+t\mat{Z}_1$.
            \STATE Predict position-wise public-ID logits.
            \STATE Update $\theta,\mat{U},\mat{V}$ using
            $\mathcal{L}_{\mathrm{CE}}+\mathcal{L}_{\mathrm{reg}}$.
        \ENDFOR
        \STATE Construct $p_{\mathrm{anchor}}$ from learned support coordinates
        and their local smoothing scale.
        \RETURN Frozen Stage 1, flow model, $\mat{V}$, and $p_{\mathrm{anchor}}$.
    \end{algorithmic}
\end{algorithm}

\begin{algorithm}[bthp]
    \caption{Sampling with \method{}.}
    \label{alg:uniflow_sampling}
    \begin{algorithmic}[1]
        \REQUIRE Anchor prior $p_{\mathrm{anchor}}$, basis $\mat{V}$,
        flow $\mat{V}_\theta$, branch codebooks
        $\{\mathcal{C}^{(m)}\}_{m=1}^{M}$,
        shared codebook $\mathcal{C}_{\mathrm{shared}}$, and decoder
        $\mathcal{D}_{\mathrm{dec}}$.
        \STATE Sample
        $\widetilde{\vect{u}}_0\sim p_{\mathrm{anchor}}$.
        \STATE Project through $\mat{V}^{\top}$, reshape to
        $\hat{\mat{Z}}_0\in\R^{L\times d_c}$, and apply row normalization.
        \STATE
        $\hat{\mat{Z}}_1
        \leftarrow
        \operatorname{ODESolve}
        (\mat{V}_\theta,\hat{\mat{Z}}_0,0,1)$.
        \FOR{$i=1,\ldots,L$}
            \STATE
            $\hat y_i
            \leftarrow
            \arg\max_k
            \operatorname{cos}
            (
            \hat{\vect{z}}_{1,i},
            \vect{c}^{\mathrm{shared}}_k
            )$.
            \STATE
            $\hat{\vect{h}}_i
            \leftarrow
            \operatorname{concat}
            (
            \vect{c}^{(1)}_{\hat y_i},
            \ldots,
            \vect{c}^{(M)}_{\hat y_i}
            )$.
        \ENDFOR
        \STATE
        $\hat{\mat{X}}
        \leftarrow
        \mathcal{D}_{\mathrm{dec}}(\hat{\mat{H}})$.
        \RETURN $\hat{\mat{X}}$.
    \end{algorithmic}
\end{algorithm}

	\section{Experiments}

	\label{sec:experiments}
	\subsection{Experimental Setup}
    \noindent\textbf{Benchmarks.}We evaluate on four datasets spanning diverse domains and complexities, following the protocol established by prior work~\citep{yoon2019time,chen2024sdformer}:	
	\begin{itemize}
		\item \textbf{Sines}: Synthetic sinusoidal sequences with varying frequencies and phases, 5 features. This serves as a sanity check for basic temporal pattern capture.
		\item \textbf{Stocks}: Daily Google stock prices from 2004--2019, including open/close/high/low prices and volume, 6 features. Real financial data with complex non-stationary dynamics.
		\item \textbf{ETTh}: Hourly electrical transformer temperature measurements, 7 features. Industrial time series with periodic patterns and trend components.
		\item \textbf{Energy}: UCI appliance energy use dataset with weather and environmental sensors, 28 features. High-dimensional multivariate series with complex cross-correlations.
	\end{itemize}

    \noindent\textbf{Baselines.} We compare \method{} with representative time-series generation
    methods across five generative paradigms.
    \textbf{GAN-based methods} include TimeGAN~\citep{yoon2019time},
    which combines adversarial and supervised learning to preserve
    temporal dynamics in a learned embedding space, and
    COT-GAN~\citep{xu2020cot}, which employs causal optimal
    transport to capture temporal dependencies in sequential data.
    \textbf{VAE-based methods} include TimeVAE~\citep{desai2021timevae},
    which learns probabilistic latent representations and supports
    interpretable temporal components such as trends and seasonality.
    \textbf{Diffusion-based methods} include
    DiffWave~\citep{kong2021diffwave}, which generates waveforms through
    iterative reverse diffusion;
    DiffTime~\citep{coletta2024constrained}, which adapts diffusion
    modeling to time-series generation and supports constraint-aware
    sampling; and
    Diffusion-TS~\citep{yuan2024diffusion}, which incorporates
    disentangled temporal representations and Fourier-based objectives
    for interpretable sequence generation.
    \textbf{Discrete-token methods} include
    SDformer-ar~\citep{chen2024sdformer} and SDformer-m~\citep{chen2024sdformer}, which combine similarity-driven
    vector quantization with autoregressive or masked discrete
    Transformer modeling. We additionally include MSDformer~\citep{feng2026msdformer}, a multi-scale autoregressive discrete generative variant of SDformer, which demonstrates competitive generation performance in long-term settings.
    \textbf{Flow-based methods} include
    FlowTS~\citep{hu2025flowts}, which employs rectified flow
    to learn continuous transport trajectories for efficient
    time-series generation.

     \begin{table*}[!t]
        \centering

        \small
        \setlength{\tabcolsep}{3.0pt}
        \renewcommand{\arraystretch}{1.40}
    
        \resizebox{\textwidth}{!}{
        \begin{tabular}{@{}ll*{12}{c}@{}}
            \toprule
    
            \multirow{2}{*}{\textbf{Category}}
            & \multirow{2}{*}{\textbf{Method}}
            & \multicolumn{3}{c}{\textbf{Sine}}
            & \multicolumn{3}{c}{\textbf{Stock}}
            & \multicolumn{3}{c}{\textbf{ETTh}}
            & \multicolumn{3}{c}{\textbf{Energy}} \\
    
            \cmidrule(lr){3-5}
            \cmidrule(lr){6-8}
            \cmidrule(lr){9-11}
            \cmidrule(l){12-14}
    
            & & \textbf{DS} $\downarrow$
              & \textbf{PS} $\downarrow$
              & \textbf{C-FID} $\downarrow$
              & \textbf{DS} $\downarrow$
              & \textbf{PS} $\downarrow$
              & \textbf{C-FID} $\downarrow$
              & \textbf{DS} $\downarrow$
              & \textbf{PS} $\downarrow$
              & \textbf{C-FID} $\downarrow$
              & \textbf{DS} $\downarrow$
              & \textbf{PS} $\downarrow$
              & \textbf{C-FID} $\downarrow$ \\
    
            \midrule
            \addlinespace[2pt]
    
            \multirow{2}{*}{GAN-based}
            & TimeGAN
            & .011$\pm$.008 & \underline{.093$\pm$.019} & .101$\pm$.014
            & .102$\pm$.021 & .038$\pm$.001 & .103$\pm$.013
            & .114$\pm$.055 & .124$\pm$.001 & .300$\pm$.013
            & .236$\pm$.012 & .273$\pm$.004 & .767$\pm$.103 \\
    
            & COT-GAN
            & .254$\pm$.137 & .100$\pm$.000 & 1.337$\pm$.068
            & .230$\pm$.016 & .047$\pm$.001 & .408$\pm$.086
            & .325$\pm$.099 & .129$\pm$.000 & .980$\pm$.071
            & .498$\pm$.002 & .259$\pm$.000 & 1.039$\pm$.028 \\
    
            \addlinespace[2pt]
            \midrule
            \addlinespace[2pt]
    
            VAE-based
            & TimeVAE
            & .041$\pm$.044 & \underline{.093$\pm$.000} & .307$\pm$.060
            & .145$\pm$.120 & .039$\pm$.000 & .215$\pm$.035
            & .209$\pm$.058 & .126$\pm$.004 & .805$\pm$.186
            & .499$\pm$.000 & .292$\pm$.000 & 1.631$\pm$.142 \\
    
            \addlinespace[2pt]
            \midrule
            \addlinespace[2pt]
    
            \multirow{3}{*}{Diffusion-based}
            & DiffWave
            & .017$\pm$.008 & \underline{.093$\pm$.000} & .014$\pm$.002
            & .232$\pm$.061 & .047$\pm$.000 & .232$\pm$.032
            & .190$\pm$.008 & .130$\pm$.001 & .873$\pm$.061
            & .493$\pm$.004 & .251$\pm$.000 & 1.031$\pm$.131 \\
    
            & DiffTime
            & .013$\pm$.006 & \underline{.093$\pm$.000} & .006$\pm$.001
            & .097$\pm$.016 & .038$\pm$.001 & .236$\pm$.074
            & .100$\pm$.007 & .121$\pm$.004 & .299$\pm$.044
            & .445$\pm$.004 & .252$\pm$.000 & .279$\pm$.045 \\
    
            & Diffusion-TS
            & .006$\pm$.007 & \underline{.093$\pm$.000} & .006$\pm$.000
            & .067$\pm$.015 & \textbf{.036$\pm$.000} & .147$\pm$.025
            & .061$\pm$.009 & \underline{.119$\pm$.002} & .116$\pm$.010
            & .122$\pm$.003 & \underline{.250$\pm$.000} & .089$\pm$.024 \\
    
            \addlinespace[2pt]
            \midrule
            \addlinespace[2pt]
    
            \multirow{2}{*}{Discrete-token}
            & SDformer-m
            & .008$\pm$.004 & \underline{.093$\pm$.000} & .010$\pm$.002
            & .020$\pm$.011 & \underline{.037$\pm$.000} & .034$\pm$.008
            & .022$\pm$.001 & \underline{.119$\pm$.002} & .019$\pm$.003
            & .062$\pm$.006 & \underline{.250$\pm$.000} & .041$\pm$.005 \\
    
            & SDformer-ar
            & .006$\pm$.004 & \underline{.093$\pm$.000} & \underline{.001$\pm$.000}
            & \underline{.010$\pm$.006} & \underline{.037$\pm$.000} & \underline{.002$\pm$.000}
            & \textbf{.003$\pm$.001} & \textbf{.118$\pm$.002} & \underline{.008$\pm$.001}
            & \underline{.006$\pm$.004} & \textbf{.249$\pm$.000} & \underline{.003$\pm$.000} \\
    
            \addlinespace[2pt]
            \midrule
            \addlinespace[2pt]
    
            Flow-based
            & FlowTS
            & \underline{.005$\pm$.005} & \textbf{.092$\pm$.000} & .002$\pm$.000
            & .019$\pm$.013 & \textbf{.036$\pm$.000} & .015$\pm$.003
            & .011$\pm$.015 & \textbf{.118$\pm$.005} & .024$\pm$.001
            & .053$\pm$.010 & \underline{.250$\pm$.000} & .031$\pm$.004 \\
    
            \addlinespace[3pt]
            \midrule
            \addlinespace[3pt]
    
            \textbf{Ours}
            & \textbf{\method{}}
            & \textbf{.003$\pm$.001} & \textbf{.092$\pm$.000} & \textbf{.0004$\pm$.0000}
            & \textbf{.005$\pm$.002} & \textbf{.036$\pm$.000} & \textbf{.001$\pm$.000}
            & \textbf{.003$\pm$.001} & \textbf{.118$\pm$.001} & \textbf{.001$\pm$.000}
            & \textbf{.003$\pm$.002} & \textbf{.249$\pm$.000} & \textbf{.001$\pm$.000} \\
    
            \addlinespace[2pt]
            \midrule
            \addlinespace[2pt]
    
            Reference
            & Original
            & --- & \textit{.094$\pm$.001} & ---
            & --- & \textit{.036$\pm$.001} & ---
            & --- & \textit{.121$\pm$.005} & ---
            & --- & \textit{.250$\pm$.003} & --- \\
    
            \bottomrule
        \end{tabular}
        }
        \caption{Unconditional generation results on standard benchmarks ($\ell=24$). Best results in \textbf{bold}, second best \underline{underlined}. Lower is better for all metrics. Original denotes the real-data PS reference. Results show mean $\pm$ std over 5 runs.}
        \label{tab:main_short}
    \end{table*}

    \noindent\textbf{Evaluation Metrics.} To comprehensively evaluate the fidelity and utility of generated time series, we report three complementary metrics: Discriminative Score, Predictive Score, and Context-FID, with results averaged over five random seeds. 
    
    To be specific, \textbf{Discriminative Score (DS)} evaluates the distinguishability between real and generated time series using a post-hoc binary classifier. Following TimeGAN and Diffusion-TS~\citep{yoon2019time,yuan2024diffusion}, we train a GRU-based classifier with a sigmoid output layer on balanced real and synthetic samples, and evaluate its accuracy on a held-out test set. The score is defined as $\mathrm{DS}=|\mathrm{Acc}-0.5|$, where $\mathrm{Acc}$ denotes the classification accuracy. A lower DS indicates that generated sequences are more difficult to distinguish from real data, with zero corresponding to chance-level discrimination. The classifier is trained using binary cross-entropy and Adam for 2,000 iterations with a batch size of 128. \textbf{Predictive Score (PS)}
    evaluates temporal predictive fidelity by training a GRU-based
    one-step-ahead predictor $g_{\mathrm{syn}}$ on synthetic sequences
    and evaluating its mean absolute error (MAE) on held-out real sequences:
    \begin{equation}
        \mathrm{PS}
        =
        \frac{1}{N}
        \sum_{i=1}^{N}
        \operatorname{MAE}
        \left(
        \bm{x}_{i,2:T},
        g_{\mathrm{syn}}(\bm{x}_{i,1:T-1})
        \right).
        \label{eq:appendix_ps}
    \end{equation}
    Here, $N$ denotes the number of real test sequences, and MAE is computed
    over all predicted time steps and variables. Lower PS indicates better
    preservation of temporal predictive structure.
    The predictor is trained using Adam with an L1 loss for 5,000 iterations
    and a batch size of 128. \textbf{Context-FID (C-FID)}
    evaluates the distributional fidelity of generated time series
    in a contextual representation space. Following~\citep{yuan2024diffusion, jeha2022psa, feng2026msdformer},
    we employ a TS2Vec encoder~\citep{yue2022ts2vec}, pretrained on
    real training data, to extract 320-dimensional embeddings from
    full-length real and synthetic sequences. The Fréchet distance
    between their Gaussian-approximated embedding distributions is
    computed as
    \begin{equation}
    \mathrm{C\mbox{-}FID}
    =\|\bm\mu_r-\bm\mu_s\|_2^2
    +\operatorname{Tr}\!\left(
    \bm\Sigma_r+\bm\Sigma_s
    -2(\bm\Sigma_r\bm\Sigma_s)^{1/2}\right),
    \label{eq:appendix_context_fid}
    \end{equation}
    where $\bm\mu_r,\bm\mu_s$ and $\bm\Sigma_r,\bm\Sigma_s$
    denote the empirical means and covariance matrices of real
    and synthetic embeddings, respectively. Lower C-FID indicates
    closer alignment between their contextual feature distributions.

    \begin{table}[bthp]
		\centering
		\small
		\resizebox{0.95\columnwidth}{!}{
			\begin{tabular}{@{}cccccccc@{}}
				\toprule
				\textbf{$\ell$} & \textbf{Branch Dims} & \textbf{Branch Depths} & \textbf{Patch Sizes} & \textbf{Shared Weights} & \textbf{$K$} & \textbf{$d_c$} & \textbf{$s$} \\
				\midrule
				24 & $192/320$ & $2/4$ & $2/4$ & $0.4/0.6$ & 512 & 512 & 4 \\
				64 & $192/320$ & $2/4$ & $2/4$ & $0.4/0.6$ & 512 & 512 & 4 \\
				128 & $192/320$ & $2/4$ & $2/4$ & $0.4/0.6$ & 512 & 512 & 4 \\
				256 & $192/320$ & $2/4$ & $2/4$ & $0.4/0.6$ & 512 & 512 & 4 \\
				\bottomrule
			\end{tabular}
		}
		\caption{Stage 1 tokenizer settings used by \method{} across sequence lengths.}
		\label{tab:app_stage1_configs}
	\end{table}

    \begin{table}[bthp]
		\centering
		\small
		\resizebox{0.95\columnwidth}{!}{
			\begin{tabular}{@{}ccccccc@{}}
				\toprule
				\textbf{$\ell$} & \textbf{Rank $r$} & \textbf{DiT Layers} & \textbf{$d_{\text{model}}$} & \textbf{Heads} & \textbf{ODE Steps} & \textbf{Bandwidth Factor} \\
				\midrule
				24 & 256 & 1 & 512 & 16 & 20 & 0.005 \\
				64 & 256--512 & 3 & 1024 & 16 & 20 & 0.005 \\
				128 & 512 & 3 & 1024 & 16 & 20 & 0.005 \\
				256 & 1024 & 3 & 1024 & 16 & 20 & 0.005 \\
				\bottomrule
			\end{tabular}
		}
		\caption{Stage 2 generator settings used by \method{} across sequence lengths.}
		\label{tab:app_stage2_configs}
	\end{table}

	\noindent\textbf{Implementation Details.} We evaluate \method{} on unconditional generation with $\ell=24$ and on longer horizons $\ell \in \{64,128,256\}$ within above four datasets. For \method{}, Stage 1 always uses a two-branch shared-index tokenizer with $K=512$, fused width $d_c=512$, branch widths $192/320$, depths $2/4$, patch sizes $2/4$, weights $0.4/0.6$, and downsampling rate $s=4$. The datails are show in Tables~\ref{tab:app_stage1_configs} and~\ref{tab:app_stage2_configs}. Also, each Stage 2 generator is paired only with its matched Stage 1 checkpoint, so the reported results do not mix incompatible tokenizers and generators. All experiments were run on a single NVIDIA GPU with 8\,GB memory, with results averaged over five runs.

	\begin{table*}[t]
		\centering
		\small
		\resizebox{0.9\textwidth}{!}{
			\begin{tabular}{@{}lccc ccc ccc@{}}
				\toprule
				& \multicolumn{3}{c}{\textbf{DS $\downarrow$}} & \multicolumn{3}{c}{\textbf{PS $\downarrow$}} & \multicolumn{3}{c}{\textbf{C-FID $\downarrow$}} \\
				\cmidrule(lr){2-4} \cmidrule(lr){5-7} \cmidrule(lr){8-10}
				\textbf{Method} & $\ell=64$ & $\ell=128$ & $\ell=256$ & $\ell=64$ & $\ell=128$ & $\ell=256$ & $\ell=64$ & $\ell=128$ & $\ell=256$ \\
				\midrule
				\multicolumn{10}{c}{\textit{ETTh Dataset}} \\
				\midrule
				TimeGAN & .227$\pm$.078 & .188$\pm$.074 & .442$\pm$.056 & .132$\pm$.008 & .153$\pm$.014 & .220$\pm$.008 & 1.130$\pm$.102 & 1.553$\pm$.169 & 5.872$\pm$.208 \\
				COT-GAN & .296$\pm$.348 & .451$\pm$.080 & .461$\pm$.010 & .135$\pm$.003 & .126$\pm$.001 & .129$\pm$.000 & 3.008$\pm$.277 & 2.639$\pm$.427 & 4.075$\pm$.894 \\
				TimeVAE & .171$\pm$.142 & .154$\pm$.087 & .178$\pm$.076 & .118$\pm$.004 & .113$\pm$.005 & .110$\pm$.027 & .827$\pm$.146 & 1.062$\pm$.134 & .826$\pm$.093 \\
				DiffWave & .254$\pm$.074 & .274$\pm$.047 & .304$\pm$.068 & .133$\pm$.008 & .129$\pm$.003 & .132$\pm$.001 & 1.543$\pm$.153 & 2.354$\pm$.170 & 2.899$\pm$.289 \\
				DiffTime & .150$\pm$.003 & .176$\pm$.015 & .243$\pm$.005 & .118$\pm$.004 & .120$\pm$.008 & .118$\pm$.003 & 1.279$\pm$.083 & 2.554$\pm$.318 & 3.524$\pm$.830 \\
				Diffusion-TS & .106$\pm$.048 & .144$\pm$.060 & .060$\pm$.030 & .116$\pm$.000 & .110$\pm$.003 & .109$\pm$.013 & .631$\pm$.058 & .787$\pm$.062 & .423$\pm$.038 \\
				FlowTS & \underline{.010$\pm$.004} & .040$\pm$.012 & .081$\pm$.022 & .115$\pm$.005 & \textbf{.104$\pm$.013} & .107$\pm$.005 & .039$\pm$.003 & .128$\pm$.007 & .302$\pm$.018 \\
				SDformer-ar & .018$\pm$.007 & .013$\pm$.005 & .008$\pm$.006 & .116$\pm$.006 & .110$\pm$.007 & \textbf{.095$\pm$.003} & .018$\pm$.003 & .024$\pm$.001 & .021$\pm$.001 \\
                MSDformer &.013$\pm$.012 &\underline{.006$\pm$.003} &\underline{.007$\pm$.005} &\underline{.112$\pm$.008} &.110$\pm$.009 & .110$\pm$.005 & \underline{.015$\pm$.002} &\underline{.023$\pm$.002} &\underline{.018$\pm$.002} \\
				\textbf{\method{}} & \textbf{.003$\pm$.000} & \textbf{.006$\pm$.002} & \textbf{.005$\pm$.002} & \textbf{.112$\pm$.003} & \underline{.108$\pm$.006} & \underline{.105$\pm$.005} & \textbf{.001$\pm$.000} & \textbf{.001$\pm$.000} & \textbf{.001$\pm$.000} \\
				\textit{Original} & --- & --- & --- & \textit{.114$\pm$.006} & \textit{.108$\pm$.005} & \textit{.106$\pm$.010} & --- & --- & --- \\
				\midrule
				\multicolumn{10}{c}{\textit{Energy Dataset}} \\
				\midrule
				TimeGAN & .498$\pm$.001 & .499$\pm$.001 & .499$\pm$.000 & .291$\pm$.003 & .303$\pm$.002 & .351$\pm$.004 & 1.230$\pm$.070 & 2.535$\pm$.372 & 5.032$\pm$.831 \\
				COT-GAN & .499$\pm$.001 & .499$\pm$.001 & .498$\pm$.004 & .262$\pm$.002 & .269$\pm$.002 & .275$\pm$.004 & 1.824$\pm$.144 & 1.822$\pm$.271 & 2.533$\pm$.467 \\
				TimeVAE & .499$\pm$.000 & .499$\pm$.000 & .499$\pm$.000 & .302$\pm$.001 & .318$\pm$.000 & .353$\pm$.003 & 2.662$\pm$.087 & 3.125$\pm$.106 & 3.768$\pm$.998 \\
				DiffWave & .497$\pm$.004 & .499$\pm$.001 & .499$\pm$.000 & .252$\pm$.001 & .252$\pm$.000 & .251$\pm$.000 & 2.697$\pm$.418 & 5.552$\pm$.528 & 5.572$\pm$.584 \\
				DiffTime & .328$\pm$.031 & .396$\pm$.024 & .437$\pm$.095 & .252$\pm$.000 & .251$\pm$.000 & .251$\pm$.000 & .762$\pm$.157 & 1.344$\pm$.131 & 4.735$\pm$.729 \\
				Diffusion-TS & .078$\pm$.021 & .143$\pm$.075 & .290$\pm$.123 & .249$\pm$.000 & .247$\pm$.001 & .245$\pm$.001 & .135$\pm$.017 & .087$\pm$.019 & .126$\pm$.024 \\
				SDformer-ar & .010$\pm$.007 & .013$\pm$.007 & .017$\pm$.003 & \underline{.247$\pm$.001} & \underline{.244$\pm$.000} & \underline{.243$\pm$.002} & .031$\pm$.002 & .036$\pm$.002 & .041$\pm$.003 \\
                MSDformer &\underline{.009$\pm$.006} &\underline{.009$\pm$.006} &\underline{.010$\pm$.009}
                &\underline{.247$\pm$.001} &\underline{.244$\pm$.000} &.243$\pm$.003 & \underline{.027$\pm$.003} &\underline{.034$\pm$.004} &\underline{.032$\pm$.002} \\
				\textbf{\method{}} & \textbf{.003$\pm$.001} & \textbf{.006$\pm$.003} & \textbf{.007$\pm$.005} & \textbf{.247$\pm$.000} & \textbf{.243$\pm$.001} & \textbf{.242$\pm$.002} & \textbf{.001$\pm$.000} & \textbf{.001$\pm$.000} & \textbf{.002$\pm$.000} \\
				\textit{Original} & --- & --- & --- & \textit{.245$\pm$.002} & \textit{.243$\pm$.000} & \textit{.243$\pm$.000} & --- & --- & --- \\
				\bottomrule
			\end{tabular}
		}
		\vspace{-1mm}
		\caption{Long-horizon generation on ETTh and Energy ($\ell=64,128,256$). Best and second-best results are \textbf{bold} and \underline{underlined}; all metrics are lower-better. Mean $\pm$ std over 5 runs; Original is the real-data PS reference.}
		\label{tab:main_long}
		\vspace{2mm}
	\end{table*}

	\subsection{Main Results}
	\label{sec:main_results}
	
	\subsubsection{Short-Sequence Generation ($\ell=24$).} 
	Table~\ref{tab:main_short} shows that \method{} is already strong in the short-horizon regime, where autoregressive error accumulation is less pronounced and several baselines remain competitive. The largest separation appears in DS and C-FID, the two metrics most directly tied to distributional realism. On Energy, for example, \method{} reduces DS from 0.006 to 0.003 and C-FID from 0.003 to 0.001 compared with the strongest token baseline. PS remains competitive across datasets. Together with the tokenizer ablation, this result shows that the shared multi-scale representation is useful even before long-horizon error accumulation becomes dominant. Also, we found that \method{}'s parallel transport also provides an efficiency advantage: in our wall-clock runs, generating 1,024 short-horizon samples is about 3--7.5$\times$ faster than autoregressive token decoding, which is shown in Table~\ref{tab:inference}.

    \begin{table}[t]
		\centering
		\small
		\resizebox{0.90\columnwidth}{!}{
			\begin{tabular}{@{}lcccc@{}}
				\toprule
				\textbf{Method} & Sine & Stock & ETTh & Energy \\
				\midrule
				Diffusion-TS & 9.57 & 10.12 & 11.85 & 34.92 \\
				SDformer-ar & 0.15 & 0.18 & 0.40 & 0.15 \\
                MSDformer & 0.95 & 0.96 & 0.30 & 0.96 \\
				\textbf{\method{} (Ours)} & \textbf{0.05} & \textbf{0.04} & \textbf{0.04} & \textbf{0.04} \\
				\midrule
				\textit{Speedup vs AR} & 3.0$\times$ & 4.5$\times$ & 7.5$\times$ & 3.8$\times$ \\
				\bottomrule
			\end{tabular}
		}
		\caption{Inference time (seconds) for generating 1,024 samples at $\ell=24$. The comparison is calculated relative to the
        fastest autoregressive baseline on each dataset.}
		\label{tab:inference}
	\end{table}

	\subsubsection{Long-Sequence Generation ($\ell \in \{64,128,256\}$).}

    Table~\ref{tab:main_long} evaluates long-horizon generation on ETTh and Energy. Across all sequence lengths, \method{} consistently achieves the lowest DS and C-FID while maintaining competitive PS, demonstrating strong distributional fidelity as the generation horizon increases. Compared with the autoregressive discrete model SDformer-ar, \method{} reduces DS from 0.008 to 0.005 and C-FID from 0.021 to 0.001 on ETTh at $\ell=256$, with corresponding reductions from 0.017 to 0.007 and 0.041 to 0.002 on Energy. Notably, MSDformer, a multi-scale autoregressive extension of SDformer, further improves long-horizon generative performance, highlighting the value of incorporating multi-scale temporal information. Nevertheless, \method{} consistently outperforms MSDformer in DS and C-FID across both datasets and all evaluated lengths, suggesting that multi-scale representations can be effectively integrated into a unified non-autoregressive generative interface. These results demonstrate the effectiveness of \method{} in modeling long-range temporal distributions without relying on sequential token generation.

    \vspace{-0.05in}

    \subsection{Cross-Scale Complementarity and Generative Consistency}
    \label{sec:single_scale_robustness}
    
    A central motivation of the shared-index design is to integrate
    complementary temporal information into a unified representation.
    To examine the contribution of cross-scale integration, we compare
    the full shared-token model with two single-scale variants that
    recover the Stage 2 trajectory using only Scale 1 or Scale 2,
    while keeping the decoder and remaining generator settings unchanged.
    
    Table~\ref{tab:single_scale_subspace} evaluates generative quality
    using DS/C-FID and MAE/MSE, together with cross-scale consistency
    through \textbf{Cycle Agree}. Given $N$ generated sequences, we re-encode each sample using
    the frozen multi-scale tokenizer and independently recover
    the public IDs from the two temporal branches, which is defined as
    \begin{equation}
    \label{eq:tokenizer_cycle_agreement}
        \text{Cycle Agree}
        =
        \frac{1}{NL}
        \sum_{n=1}^{N}
        \sum_{i=1}^{L}
        \mathbb{I}
        \left[
        \hat{y}_{i}^{(1,n)}
        =
        \hat{y}_{i}^{(2,n)}
        \right],
    \end{equation}
    where
    $\hat{y}_{i}^{(m,n)}
    =
    \arg\max_k
    \operatorname{cos}
    (\mathcal{E}^{(m)}(\hat{\mat{X}}^{(n)})_i,
    \vect{c}^{(m)}_k)$
    denotes the ID recovered independently from scale $m$.
    Higher value indicates stronger cross-scale agreement under
    the frozen tokenizer, rather than external semantic correctness.

     \begin{table}[bthp]
		\centering
		\small
		\setlength{\tabcolsep}{3.5pt}
		\resizebox{0.98\columnwidth}{!}{
			\begin{tabular}{@{}llccc@{}}
				\toprule
				\textbf{Dataset} & \textbf{Variant} & \textbf{DS / C-FID} $\downarrow$ & \textbf{MAE / MSE} $\downarrow$ & \textbf{Cycle Agree.} $\uparrow$ \\
				\midrule
				\multirow{4}{*}{Sine}
				& Real (ref.) & -- & -- & 0.665 \\
				& Shared gen. & \textbf{0.003 / 0.0004} & 0.1201 / 0.0268 & \textbf{0.766} \\
				& Scale-1 gen. & 0.008 / 0.009 & \textbf{0.1062 / 0.0208} & 0.499 \\
				& Scale-2 gen. & 0.011 / 0.007 & 0.1110 / 0.0227 & 0.540 \\
				\midrule
				\multirow{4}{*}{ETTh}
				& Real (ref.) & -- & -- & 0.711 \\
				& Shared gen. & \textbf{0.003 / 0.001} & 0.1543 / 0.0428 & \textbf{0.768} \\
				& Scale-1 gen. & 0.108 / 0.008 & \textbf{0.1457 / 0.0383} & 0.467 \\
				& Scale-2 gen. & 0.069 / 0.010 & 0.1497 / 0.0400 & 0.477 \\
				\midrule
				\multirow{4}{*}{Energy}
				& Real (ref.) & -- & -- & 0.778 \\
				& Shared gen. & \textbf{0.003 / 0.001} & 0.2011 / 0.0703 & \textbf{0.789} \\
				& Scale-1 gen. & 0.345 / 0.363 & 0.1768 / 0.0561 & 0.483 \\
				& Scale-2 gen. & 0.316 / 0.326 & \textbf{0.1704 / 0.0517} & 0.466 \\
				\bottomrule
			\end{tabular}
		}
		\vspace{-0.5mm}
		\caption{Single-view generation and tokenizer-cycle consistency. Cycle Agree. compares scale-specific IDs after decoding and re-encoding.}
		\label{tab:single_scale_subspace}
		\vspace{-0.5mm}
	\end{table}
    
    Across all three datasets, shared generation consistently achieves
    lower DS/C-FID and higher agreement than either single-scale variant.
    Although individual scales occasionally yield lower MAE/MSE,
    these improvements do not translate into better distributional
    fidelity or cross-scale consistency.
    These observations suggest that different temporal views provide
    complementary information and that shared-index modeling
    effectively integrates this information into a more coherent
    generative representation.


    \noindent\textbf{Coarse-View Distributional Consistency.}
    To further assess whether the generated sequences
    preserve temporal structure beyond their original
    resolution, we evaluate their distributions under
    non-overlapping temporal aggregation. For each window
    $w\in\{2,4\}$, we construct summary features from
    the aggregated trajectories, first-order differences,
    and channel-wise means and standard deviations.
    We then compute a Fréchet-type distance between real
    and generated features, standardized using real-data
    statistics, and average the distances across windows.

	\begin{table}[bthp]
		\centering
		\small
		\resizebox{1.0\columnwidth}{!}{
			\begin{tabular}{@{}lccccc@{}}
				\toprule
				\textbf{Dataset} & \textbf{Shared Dist.} & \textbf{S1 Dist.} & \textbf{S2 Dist.} & \textbf{S1/Shared} & \textbf{S2/Shared} \\
				\midrule
				Sine & \textbf{0.455} & 15.185 & 13.908 & 33.3$\times$ & 30.5$\times$ \\
				ETTh & \textbf{3.746} & 14.949 & 14.848 & 4.0$\times$ & 4.0$\times$ \\
				Energy & \textbf{62.493} & 369.868 & 241.368 & 5.9$\times$ & 3.9$\times$ \\
				\bottomrule
			\end{tabular}
		}
		\caption{Coarse-view distributional fidelity under non-overlapping temporal aggregation $w\in\{2,4\}$. Distances are averaged across windows; lower is better. Ratios indicate the relative distances of single-scale variants to shared generation.}
		\label{tab:app_coarse_view_consistency}
	\end{table}

\begin{figure}[bthp]
		\centering
		\includegraphics[width=0.95\columnwidth]{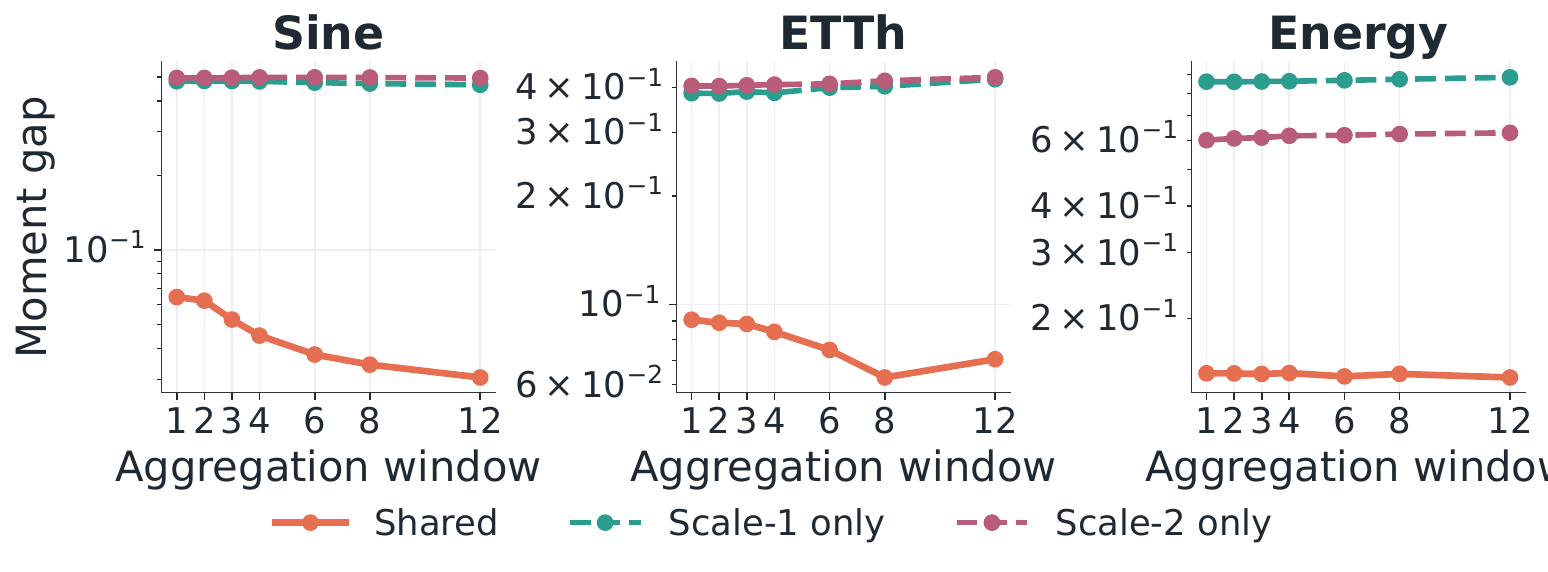}
		\\[-0.6em]
		\includegraphics[width=\columnwidth]{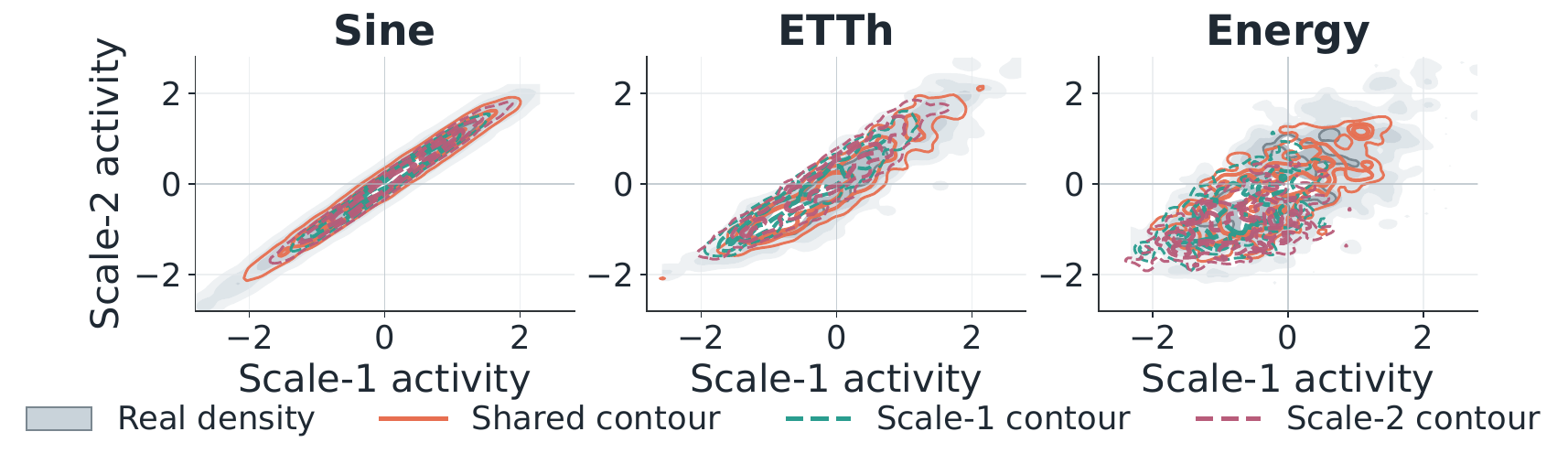}
		\\
		\includegraphics[width=\columnwidth]{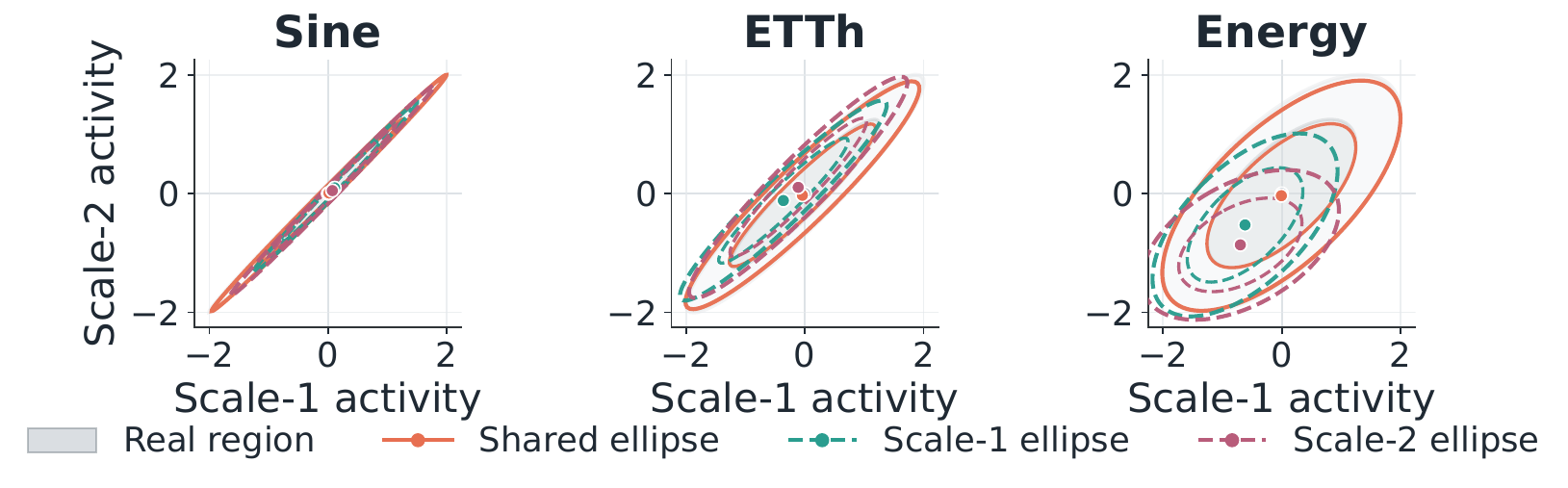}
		\vspace{-7mm}
        \caption{Multi-scale temporal structure under shared and single-scale generation.
        Top: Temporal aggregation gaps.
        Middle: Joint fine- and coarse-scale activity distributions.
        Bottom: Covariance ellipses of joint variation.
        Shared generation better preserves the real-data multi-scale structure.}
		\label{fig:new_multiscale_probes}
		\vspace{3mm}
	\end{figure}

    As shown in Table~\ref{tab:app_coarse_view_consistency},
    shared generation consistently achieves the lowest
    coarse-view distances across all three datasets.
    Compared with shared generation, the Scale-1 and
    Scale-2 variants exhibit approximately
    $30.5$--$33.3\times$ larger distances on Sine,
    $4.0\times$ on ETTh, and $3.9$--$5.9\times$ on Energy.
    These results complement the scale-aware visualizations
    in Fig.~\ref{fig:new_multiscale_probes}, showing that
    the benefits of shared-index generation extend beyond
    discrete semantic agreement to the preservation of
    observable temporal statistics under aggregation.

    Together, these analyses provide complementary evidence at both semantic and observable levels: shared-index generation improves cross-scale agreement in the learned token space while more faithfully reproducing multi-scale temporal statistics in the generated sequences. This distinction highlights the role of the unified public-ID interface in preserving information across temporal resolutions without introducing separate generative token streams.

    \begin{table}[h]
		\centering
		\small
		\resizebox{1.0\columnwidth}{!}{
			\begin{tabular}{@{}lcccc@{}}
				\toprule
				\textbf{Condition} & \textbf{Initial Acc.} & \textbf{Refined Acc.} & \textbf{Correction} & \textbf{DS / C-FID} \\
				\midrule
				Scale-1 IDs & 0.7873 & 0.9999 & 0.9996 & 0.0116 / 0.00024 \\
				Scale-2 IDs & 0.8540 & 0.9999 & 0.9997 & 0.0109 / 0.00026 \\
				\bottomrule
			\end{tabular}
		}
		\caption{Sine conditional public-ID refinement diagnostic. Initial accuracy is the token match between a single branch's recovered IDs and the jointly voted shared IDs; refined accuracy is measured after the conditional Stage 2 refinement.}
		\label{tab:app_conditional_refinement}
	\end{table}


    \begin{table*}[t]
        \centering
        \small
        \setlength{\tabcolsep}{6pt}
        \renewcommand{\arraystretch}{1.2}
        \resizebox{0.95\textwidth}{!}{
            \begin{tabular}{@{}lccccc|ccccc@{}}
                \toprule
                \multirow{2}{*}{\textbf{Dataset}}
                & \multicolumn{5}{c|}{\textbf{MSE} $\downarrow$}
                & \multicolumn{5}{c}{\textbf{MAE} $\downarrow$} \\
                \cmidrule(lr){2-6}
                \cmidrule(l){7-11}
                & \textbf{Shared}
                & \textbf{Scale-1}
                & \textbf{Scale-2}
                & \textbf{S1 $\Delta$}
                & \textbf{S2 $\Delta$}
                & \textbf{Shared}
                & \textbf{Scale-1}
                & \textbf{Scale-2}
                & \textbf{S1 $\Delta$}
                & \textbf{S2 $\Delta$} \\
                \midrule
                Sine
                & \textbf{0.00004929}
                & 0.00005341
                & 0.00006096
                & 0.00000412
                & 0.00001166
                & \textbf{0.005348}
                & 0.005533
                & 0.005819
                & 0.000185
                & 0.000471 \\
    
                Stock
                & \textbf{0.00018091}
                & 0.00018585
                & 0.00018173
                & 0.00000495
                & 0.00000082
                & \textbf{0.008570}
                & 0.008625
                & 0.008587
                & 0.000055
                & 0.000017 \\
    
                ETTh
                & \textbf{0.00010566}
                & 0.00011746
                & 0.00013094
                & 0.00001180
                & 0.00002528
                & \textbf{0.007804}
                & 0.008116
                & 0.008494
                & 0.000312
                & 0.000690 \\
    
                Energy
                & \textbf{0.00005832}
                & 0.00008911
                & 0.00016054
                & 0.00003078
                & 0.00010222
                & \textbf{0.005546}
                & 0.005986
                & 0.006699
                & 0.000440
                & 0.001153 \\
    
                \midrule
                \textbf{Avg. $\Delta$}
                & -- & -- & --
                & 0.00001291
                & 0.00003500
                & -- & -- & --
                & 0.000248
                & 0.000583 \\
                \bottomrule
            \end{tabular}
        }
        \caption{Reconstruction performance under shared and single-view
        public-ID assignments. $\Delta$ denotes the increase in reconstruction
        error relative to the shared mapping. Lower values indicate better
        reconstruction quality.}
        \label{tab:shared_id_reconstruction}
    \end{table*}

    \begin{figure}[t]
        \centering
        \includegraphics[width=\columnwidth]{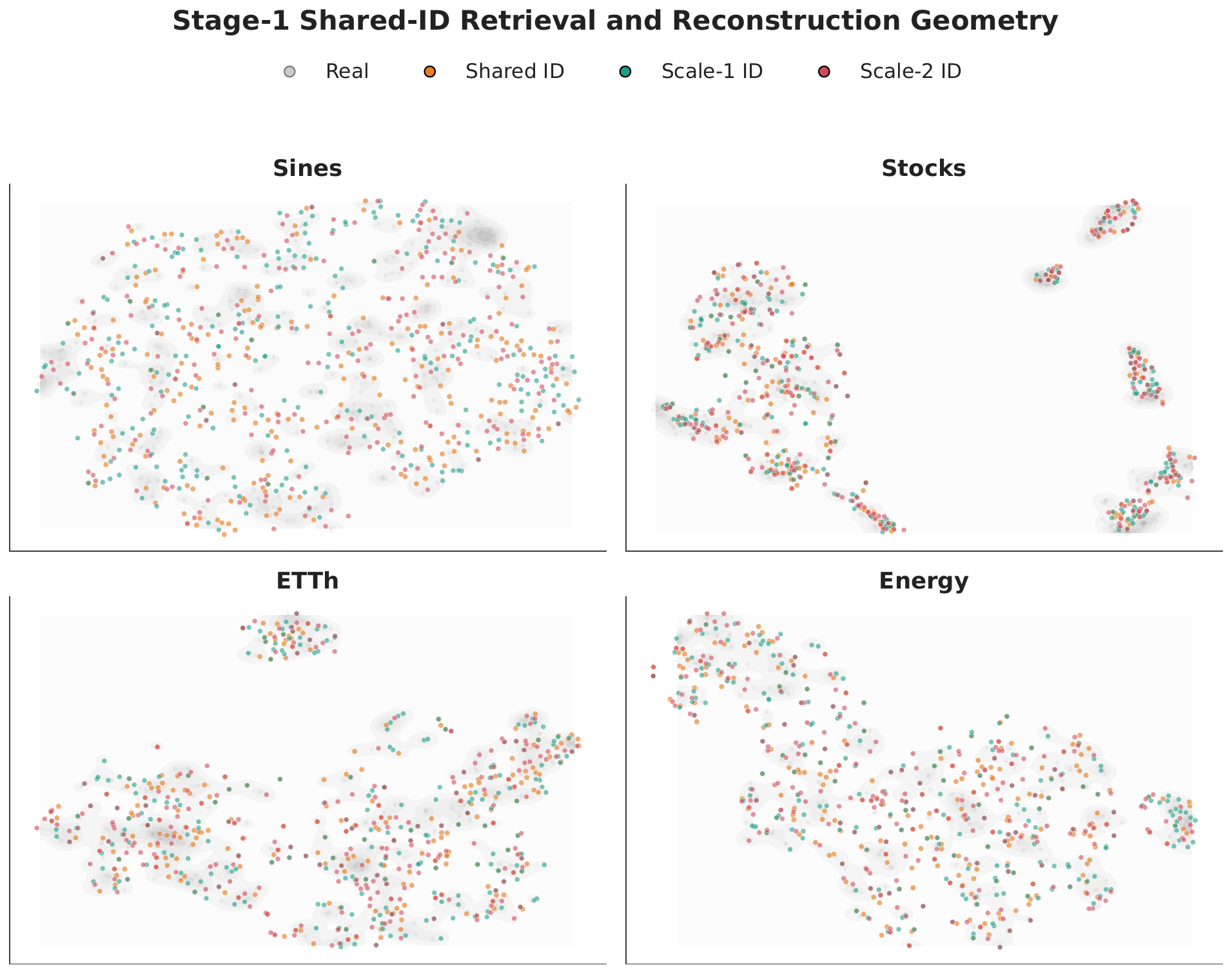}
        \caption{t-SNE visualization of real and Stage 1
        reconstructed samples under different public-ID
        assignments.}
        \label{fig:app_shared_id_tsne}
    \end{figure}

    \subsection{Effects of Shared-Index Voting on Reconstruction Fidelity}
    \label{sec:shared_reconstruction}

    A key question in shared-semantic tokenization is whether unifying multiple temporal views into a single public ID compromises reconstruction fidelity. To investigate this, we evaluate Stage 1 reconstruction by comparing the original shared-index assignment with single-view alternatives that recover public IDs using only Scale 1 or Scale 2, while keeping the learned scale-specific codebooks and decoder fixed.

    Table~\ref{tab:shared_id_reconstruction} reports reconstruction MSE/MAE and their changes relative to the shared assignment. The shared mapping consistently achieves the lowest reconstruction errors across all four datasets, indicating that aggregating cross-scale evidence produces more informative code assignments than relying on either individual view. The varying reconstruction degradation across datasets and scales highlights their data-dependent contributions, further motivating the integration of complementary temporal evidence through shared-index voting.
    
    These findings complement the single-view recoverability analysis by showing that shared-index voting improves not only discrete ID agreement but also reconstruction fidelity. Together with the Stage 2 generation results DS/C-FID in Table~\ref{tab:single_scale_subspace}, they provide evidence that the unified public-token interface supports both faithful reconstruction and high-quality generative modeling. The t-SNE visualization in Fig.~\ref{fig:app_shared_id_tsne} further illustrates the distributional overlap between real samples and Stage 1 reconstructions under different public-ID assignments.
    
    
    

    \vspace{-0.15in}

    \subsection{Single-View Conditioned Shared-ID Refinement}
    \label{sec:conditional_refinement}
    
    Building upon the single-view recovery property in
    Proposition~\ref{prop:single_view_recovery_main},
    we further investigate whether Stage 2 can refine
    incomplete public-ID assignments obtained from an
    individual temporal scale. While the proposition
    characterizes when branch-specific evidence is
    sufficient for stable ID recovery, it does not
    guarantee correct assignments at positions with
    insufficient semantic margins.
    
    To examine this setting, we conduct a controlled
    experiment on Sine, where IDs independently recovered
    from Scale 1 or Scale 2 initialize the refinement
    process and condition Stage 2 through adaptive layer
    normalization (AdaLN). The jointly voted shared IDs
    serve as the reference for evaluating token recovery.
    
    As shown in Table~\ref{tab:app_conditional_refinement},
    conditional refinement improves token accuracy from
    $78.73\%/85.40\%$ to $99.99\%$ for Scale 1/Scale 2,
    correcting over $99.9\%$ of initially mismatched IDs.
    The refined sequences also achieve DS/C-FID values
    of $0.0116/0.00024$ and $0.0109/0.00026$,
    respectively.
    
    These results extend the single-view recoverability
    analysis by showing that, in this controlled setting,
    Stage 2 can substantially reduce ambiguities remaining
    in individual-scale ID assignments. This further
    illustrates the potential of the shared-semantic
    interface for conditional cross-scale generation.

    \subsection{Ablation Study}
    \label{sec:ablation}
    
    Table~\ref{tab:ablation_energy} examines three key design
    choices of \method{}: multi-scale tokenization, the shared
    public-ID interface, and flow initialization of Energy dataset.
    
    \noindent\textbf{Multi-scale Tokenization.}
    Introducing multiple temporal scales alone does not
    guarantee better generation. The dilation-only variant
    performs worse than single-scale token flow
    (DS: $0.027$ vs. $0.006$), whereas patch-based
    multi-scale tokenization achieves substantially lower
    DS/C-FID $0.003/0.001$. This contrast indicates
    that the effectiveness of multi-scale modeling depends
    on how temporal information is represented and aligned,
    rather than simply increasing the number of scales.
    
    \noindent\textbf{Shared Semantic Interface.}
    Keeping the Stage 2 architecture and token sequence
    length fixed, replacing jointly voted public IDs with
    Scale 1 or Scale 2 IDs substantially degrades
    DS/C-FID, from $0.003/0.001$ to $0.345/0.363$
    and $0.316/0.326$, respectively.
    This result highlights the importance of integrating
    cross-scale evidence into a unified semantic identity,
    rather than relying on individual scale-specific
    assignments.
    
    \noindent\textbf{Geometry-aware Initialization.}
    Replacing the learned anchor prior with Gaussian
    coordinates increases DS/C-FID to $0.218/0.279$.
    A fitted-mixture prior improves performance to
    $0.047/0.031$, but remains inferior to the learned
    anchor prior $0.003/0.001$.
    These results suggest that initialization informed
    by the learned low-rank support structure provides
    a more suitable starting distribution for categorical
    flow transport than the evaluated alternatives.
    
    Overall, the three ablations reveal complementary
    roles across the two stages: multi-scale tokenization
    determines the temporal information captured by
    the tokenizer, shared-index voting consolidates this
    information into a unified generative interface,
    and the anchor prior determines the initialization
    geometry for subsequent transport.
    Their combined effectiveness supports the
    representation-to-generation design of \method{}.

	
	\begin{table}[t]
		\centering
		\small
		\resizebox{0.95\columnwidth}{!}{
			\begin{tabular}{@{}llcc@{}}
				\toprule
				\textbf{Component} & \textbf{Variant} & \textbf{DS} $\downarrow$ & \textbf{C-FID} $\downarrow$ \\
				\midrule
				\multirow{3}{*}{Tokenizer}
				& Single-scale token flow & 0.006 & 0.003 \\
				& Multi-scale, dilation-only & 0.027 & 0.018 \\
				& Multi-scale, patch-based & \textbf{0.003} & \textbf{0.001} \\
				\midrule
				\multirow{3}{*}{Interface}
				& Shared public IDs & \textbf{0.003} & \textbf{0.001} \\
				& Scale-1 IDs only & 0.345 & 0.363 \\
				& Scale-2 IDs only & 0.316 & 0.326 \\
				\midrule
				\multirow{3}{*}{Prior}
				& Gaussian coordinate prior & 0.218 & 0.279 \\
				& Fitted-mixture prior & 0.047 & 0.031 \\
				& Learned anchor prior & \textbf{0.003} & \textbf{0.001} \\
				\bottomrule
			\end{tabular}
		}
		\vspace{-0.5mm}
		\caption{Ablation of multi-scale tokenization, shared-ID
interface, and flow initialization on Energy.}
		\label{tab:ablation_energy}
		\vspace{-0.5mm}
	\end{table}

    \section{Conclusion}
    \label{sec:conclusion}
    
    We presented \method{}, a unified framework for
    multi-scale time series generation that reconciles
    rich temporal representations with efficient generative
    modeling. Through shared-index voting and scale-specific
    codebooks, \method{} integrates complementary temporal
    information into a single public-ID sequence while
    preserving scale-dependent reconstruction details.
    Building on this unified representation, geometry-aware
    categorical flow matching leverages learned low-rank
    anchors to initialize continuous transport and enables
    non-autoregressive generation without sequential
    token decoding.
    
    Extensive experiments demonstrate that \method{} achieves 
    improved generative fidelity and inference efficiency across diverse benchmarks and generation horizons. 
    Further analyses validate its effectiveness in preserving reconstruction fidelity, 
    cross-scale semantic consistency, and multi-scale temporal structure. 
    These findings highlight that rich multi-scale information can be 
    integrated into a unified generative interface, enabling high-quality and efficient 
    time series generation without multiple token streams.

    \noindent\textbf{Limitations and Future Directions.} 
    Modeling temporal information across multiple scales is
    essential for capturing the complex dynamics of time series data. The current tokenization of \method{} relies on predefined
    temporal scales. Exploring adaptive scale selection and
    integration could further enhance its ability.
\bibliographystyle{IEEEtran}
\bibliography{reference.bib} 










\section{Biography Section}
\vspace{-0.5in}
\begin{IEEEbiography}[{\includegraphics[width=1in, height=1.25in,clip,keepaspectratio]{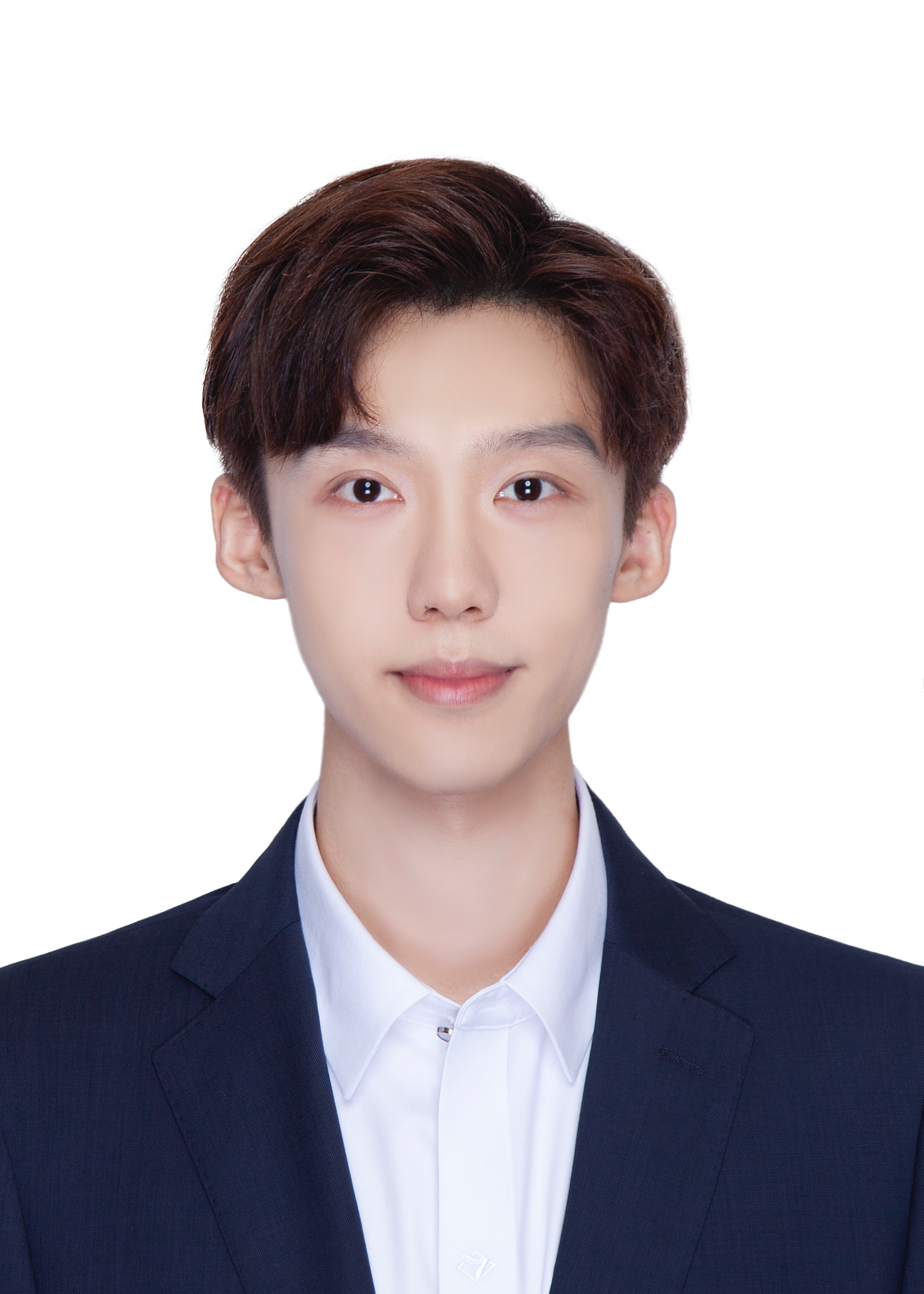}}]{Shibo Feng} is currently a researcher at Nanyang Technological University (NTU), Singapore. His research focuses on Generative AI and its applications, including large models and time series analysis. He has served as the reviewer for top-tier venues such as ICML, ICLR, NeurIPS, and TPAMI. He received his PhD’s degree in College of Computing and Data Science (CCDS) from Nanyang Technological University.
\vspace{9mm}
\end{IEEEbiography}

\begin{IEEEbiography}[{\includegraphics[width=1in,height=1.25in,clip,keepaspectratio]{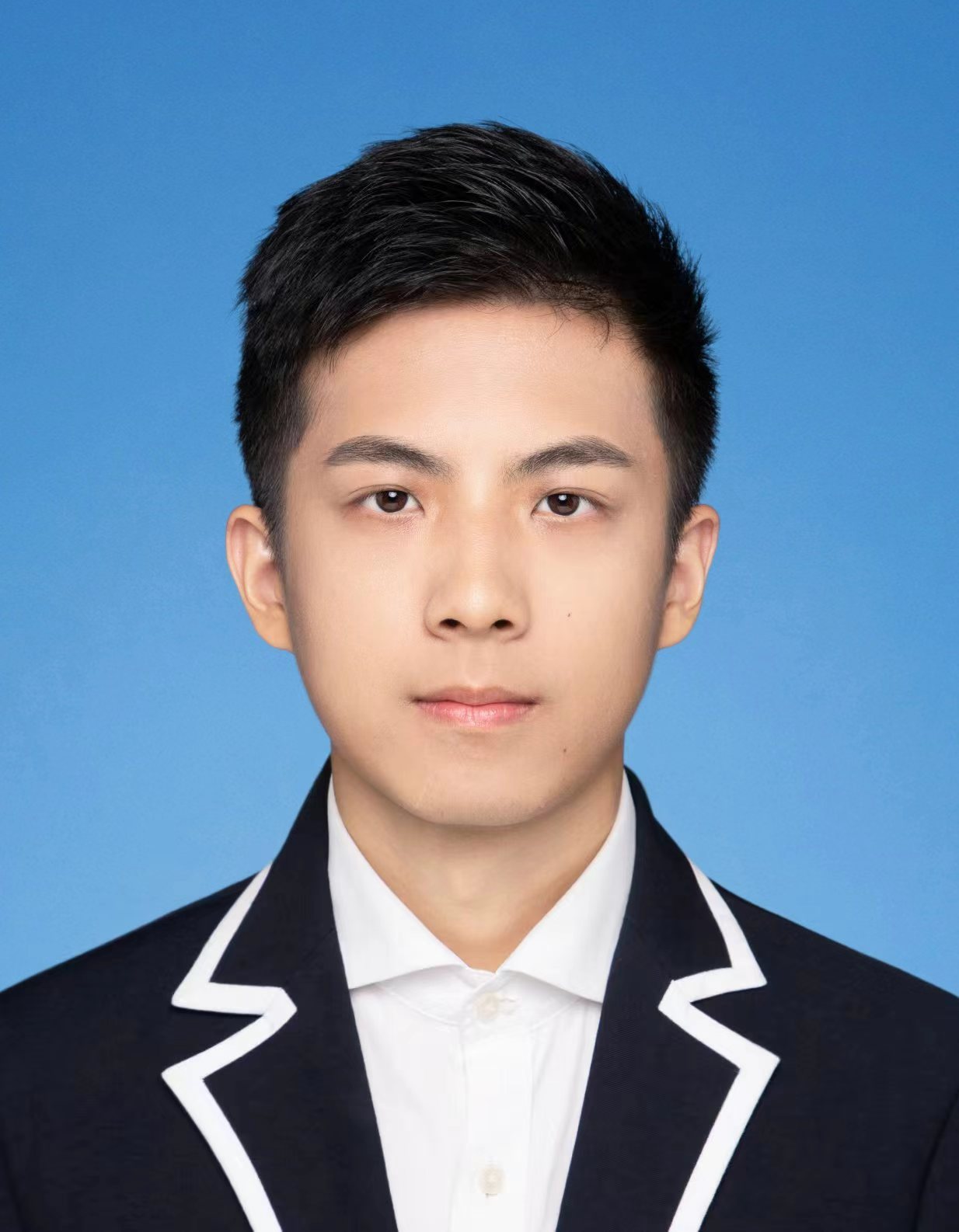}}]{Wei Li}
studies computer science and technology at Shanghai University, China (2023--2027), and has been admitted to the joint Ph.D. program of Shanghai Jiao Tong University and Beijing Zhongguancun Academy, starting in 2027. His research focuses on machine learning and data mining, including time-series generation, forecasting, clustering, and representation learning. His representative works include SDFlow (NeurIPS 2026) and APCL (KDD 2026). He has been invited to review for venues including Neural Networks, NeurIPS (ethics review), AAAI, and ICASSP. He received the National Scholarship and the Baosteel Scholarship in 2026.
\end{IEEEbiography}

\begin{IEEEbiography}[{\includegraphics[width=1in, height=1.25in,clip,keepaspectratio]{
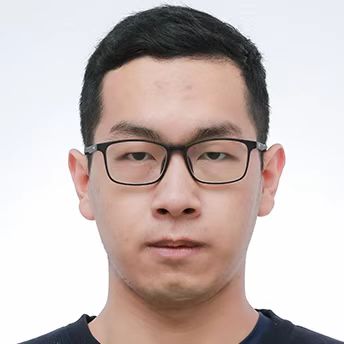}}]{Songhua Liu} is currently an Assistant Professor and Ph.D. Advisor at the School of Artificial Intelligence, Shanghai Jiao Tong University. His research focuses on generative AI, particularly efficient deep learning driven by synthetic data. He received his Ph.D. in Electronic and Computer Engineering from the National University of Singapore in 2025, under the supervision of Prof. Xinchao Wang, and his bachelor’s degree in Computer Science and Technology from Nanjing University in 2021. He previously conducted research at Rutgers University and Baidu Inc., under the supervision of Prof. Hao Wang and Tianwei Lin, respectively.
\end{IEEEbiography}

\begin{IEEEbiography}[{\includegraphics[width=1in, height=1.25in,clip,keepaspectratio]{
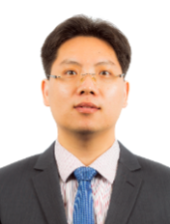}}]{Min Wu} is a Chapter Head at the A*STAR Institute of Advanced Intelligence and Computing (A*STAR IAIC), and Deputy Head of Tech Development at the AI Centre of Excellence for Manufacturing (AIMfg). He received his Ph.D. in Computer Science from Nanyang Technological University (NTU) in 2011 and his B.E. in Computer Science from the University of Science and Technology of China (USTC) in 2006. He has won several best paper awards and serves as an Associate Editor for journals including Neural Networks and IEEE Transactions on Neural Networks and Learning Systems. His current research interests focus on AI for time series data, graph data, and biological and healthcare data.
\end{IEEEbiography}

\begin{IEEEbiography}[{\includegraphics[width=1in,height=1.25in,clip,keepaspectratio]{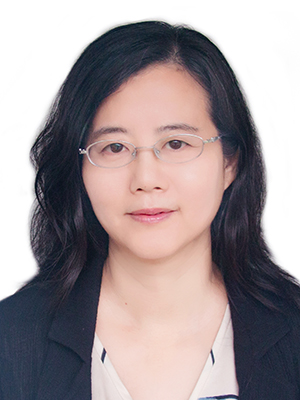}}]{Chunyan Miao} is a President’s Chair Professor and the Chair of College of Computing and Data Science at Nanyang Technological University (NTU), Singapore. She received her Ph.D. degree from NTU and was an NSERC Postdoctoral Fellow at Simon Fraser University (SFU), Canada. She was a founding faculty member of the Centre for Digital Media established by the University of British Columbia (UBC) and SFU. She was also a Tan Chin Tuan Engineering Fellow at Harvard and MIT. Dr. Miao has focused on her research in Artificial intelligence (AI) and real-world AI applications for impactful research in health, aging, education, and smart services.  She is the Founding Director of the Joint NTU-UBC Research Centre of Excellence in Active Living for the Elderly (LILY), Singapore’s first center focusing on AI-empowered solutions to population aging challenges. She is the Founding Director of the Alibaba-NTU Singapore Joint Research Institute (JRI), Alibaba’s first and largest JRI outside China. She is an Editor/Associate Editor of leading international journals including IJIT, IEEE Big Data, IEEE IoT, IEEE Access and IEEE Service Computing, and has served as Chair/TPC member of international conferences such as IEEE ICA, ICAA, ACM KDD. She serves on various national committees and is the Chair of the SCS AI Ethics Review Committee.
\vspace{9mm}
\end{IEEEbiography}

\begin{IEEEbiography}[{\includegraphics[width=1in, height=1.25in,clip,keepaspectratio]{
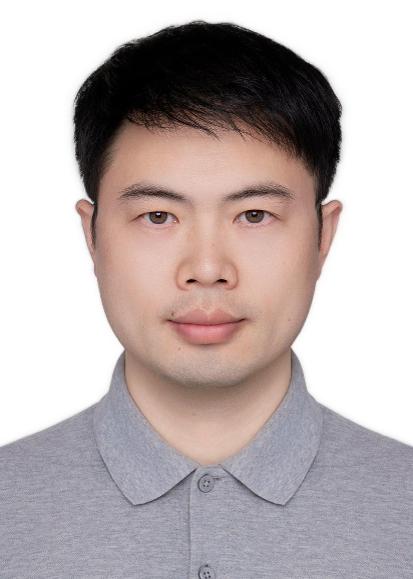}}]{Peilin Zhao} is currently a professor at Shanghai Jiao Tong University. He previously worked at Tencent Group, Ant Group, A*STAR Singapore, and Rutgers University. His research focuses on machine learning and its applications, including large models, AI for science, AI for games, and recommendation systems. He has served as an Area Chair and Associate Editor for top-tier venues such as ICML, NeurIPS, and TPAMI. He received his bachelor’s degree in mathematics from Zhejiang University and his PhD in computer science from Nanyang Technological University.
\vspace{100 mm}
\end{IEEEbiography}

 



    \clearpage

    \setcounter{section}{0}
    \begin{center}
        {\LARGE \bf Supplementary Materials for UniFlow}\\[4pt]
    \end{center}
    
	
	\section{Probabilistic Robustness of the Shared Mapping}
	\label{app:shared_margin}
	
	We analyze how weighted cross-scale evidence affects the probability of an incorrect public-ID assignment. Fix a temporal position $i$, let $y_i^\star\in\{0,\ldots,K-1\}$ denote the reference public ID, and define the branch-wise margin against a competing code $k\neq y_i^\star$ as
	\begin{equation}
		\Delta_{i,k}^{(m)}
		:=
		s_{i,y_i^\star}^{(m)}-s_{i,k}^{(m)}.
	\end{equation}
	
	\begin{theorem}[Probabilistic robustness of shared-index voting]
		\label{thm:shared_vote_robustness}
		For each competitor $k\neq y_i^\star$, assume that the margin vector
		\begin{equation}
			\boldsymbol{\Delta}_{i,k}
			:=
			\bigl(\Delta_{i,k}^{(1)},\ldots,\Delta_{i,k}^{(M)}\bigr)^\top
			=
			\boldsymbol{\mu}_{i,k}+\boldsymbol{\varepsilon}_{i,k},
		\end{equation}
		where $\mathbb{E}[\boldsymbol{\varepsilon}_{i,k}]=\mathbf{0}$ and $\boldsymbol{\varepsilon}_{i,k}$ is sub-Gaussian with covariance proxy $\boldsymbol{\Sigma}_{i,k}\succeq0$, i.e.,
		\begin{equation}
			\mathbb{E}\!\left[
			\exp\!\left(\lambda\mathbf{a}^{\top}\boldsymbol{\varepsilon}_{i,k}\right)
			\right]
			\leq
			\exp\!\left(
			\frac{\lambda^2}{2}
			\mathbf{a}^{\top}\boldsymbol{\Sigma}_{i,k}\mathbf{a}
			\right)
		\end{equation}
		for every $\lambda\in\mathbb{R}$ and $\mathbf{a}\in\mathbb{R}^{M}$. Let $\boldsymbol{\alpha}=(\alpha_1,\ldots,\alpha_M)^\top$, where $\alpha_m\geq0$ and $\sum_m\alpha_m=1$, and define
		\begin{equation}
			\widetilde{y}_i
			=
			\arg\max_{j\in\{0,\ldots,K-1\}}
			\sum_{m=1}^{M}\alpha_m s_{i,j}^{(m)}.
		\end{equation}
		If $\boldsymbol{\alpha}^{\top}\boldsymbol{\mu}_{i,k}>0$ for every $k\neq y_i^\star$, then
		\begin{equation}
			\label{eq:shared_vote_error_bound}
			\mathbb{P}(\widetilde{y}_i\neq y_i^\star)
			\leq
			\sum_{k\neq y_i^\star}
			\exp\!\left(
			-\frac{(\boldsymbol{\alpha}^{\top}\boldsymbol{\mu}_{i,k})^2}
			{2\boldsymbol{\alpha}^{\top}\boldsymbol{\Sigma}_{i,k}\boldsymbol{\alpha}}
			\right).
		\end{equation}
		When $\boldsymbol{\alpha}^{\top}\boldsymbol{\Sigma}_{i,k}\boldsymbol{\alpha}=0$ and $\boldsymbol{\alpha}^{\top}\boldsymbol{\mu}_{i,k}>0$, the corresponding term is interpreted as zero.
	\end{theorem}
	
	\begin{proof}
		For a competitor $k\neq y_i^\star$, the shared vote can prefer $k$ to $y_i^\star$ only if
		\begin{equation}
			\boldsymbol{\alpha}^{\top}\boldsymbol{\Delta}_{i,k}\leq0.
		\end{equation}
		Using $\boldsymbol{\Delta}_{i,k}=\boldsymbol{\mu}_{i,k}+\boldsymbol{\varepsilon}_{i,k}$, this event is equivalent to
		\begin{equation}
			\boldsymbol{\alpha}^{\top}\boldsymbol{\varepsilon}_{i,k}
			\leq
			-\boldsymbol{\alpha}^{\top}\boldsymbol{\mu}_{i,k}.
		\end{equation}
		The sub-Gaussian assumption implies that $\boldsymbol{\alpha}^{\top}\boldsymbol{\varepsilon}_{i,k}$ has variance proxy $\boldsymbol{\alpha}^{\top}\boldsymbol{\Sigma}_{i,k}\boldsymbol{\alpha}$. Therefore,
		\begin{equation}
			\mathbb{P}\!\left(\boldsymbol{\alpha}^{\top}\boldsymbol{\Delta}_{i,k}\leq0\right)
			\leq
			\exp\!\left(
			-\frac{(\boldsymbol{\alpha}^{\top}\boldsymbol{\mu}_{i,k})^2}
			{2\boldsymbol{\alpha}^{\top}\boldsymbol{\Sigma}_{i,k}\boldsymbol{\alpha}}
			\right).
		\end{equation}
		A prediction error requires at least one competitor to attain a score no smaller than the reference code. Applying the union bound over all $k\neq y_i^\star$ proves Equation~\eqref{eq:shared_vote_error_bound}.
	\end{proof}
	
	\begin{corollary}[Pairwise optimal voting weights]
		\label{cor:optimal_shared_vote_weights}
		Fix $i$ and $k\neq y_i^\star$, and suppose $\boldsymbol{\Sigma}_{i,k}\succ0$. Among all nonzero weight vectors, the exponent in Theorem~\ref{thm:shared_vote_robustness} is maximized by
		\begin{equation}
			\boldsymbol{\alpha}_{i,k}^{\star}
			\propto
			\boldsymbol{\Sigma}_{i,k}^{-1}\boldsymbol{\mu}_{i,k}.
		\end{equation}
		If $\boldsymbol{\Sigma}_{i,k}^{-1}\boldsymbol{\mu}_{i,k}$ is component-wise nonnegative, the simplex-constrained solution is
		\begin{equation}
			\boldsymbol{\alpha}_{i,k}^{\star}
			=
			\frac{\boldsymbol{\Sigma}_{i,k}^{-1}\boldsymbol{\mu}_{i,k}}
			{\mathbf{1}^{\top}\boldsymbol{\Sigma}_{i,k}^{-1}\boldsymbol{\mu}_{i,k}}.
		\end{equation}
	\end{corollary}
	
	\begin{proof}
		For a fixed competitor, maximizing the exponent is equivalent to maximizing
		\begin{equation}
			\mathcal{R}_{i,k}(\boldsymbol{\alpha})
			:=
			\frac{(\boldsymbol{\alpha}^{\top}\boldsymbol{\mu}_{i,k})^2}
			{\boldsymbol{\alpha}^{\top}\boldsymbol{\Sigma}_{i,k}\boldsymbol{\alpha}}.
		\end{equation}
		Let $\mathbf{b}=\boldsymbol{\Sigma}_{i,k}^{1/2}\boldsymbol{\alpha}$ and $\mathbf{c}=\boldsymbol{\Sigma}_{i,k}^{-1/2}\boldsymbol{\mu}_{i,k}$. By Cauchy--Schwarz,
		\begin{equation}
			\mathcal{R}_{i,k}(\boldsymbol{\alpha})
			=
			\frac{(\mathbf{b}^{\top}\mathbf{c})^2}{\|\mathbf{b}\|_2^2}
			\leq
			\|\mathbf{c}\|_2^2,
		\end{equation}
		with equality if and only if $\mathbf{b}\propto\mathbf{c}$, or equivalently $\boldsymbol{\alpha}\propto\boldsymbol{\Sigma}_{i,k}^{-1}\boldsymbol{\mu}_{i,k}$. Normalization to unit sum gives the simplex solution when the vector is component-wise nonnegative.
	\end{proof}
	
	\begin{remark}[Independent branch noise]
		If branch noises are independent and
		\begin{equation}
			\boldsymbol{\Sigma}_{i,k}
			=
			\operatorname{diag}(\sigma_{i,k,1}^2,\ldots,\sigma_{i,k,M}^2),
		\end{equation}
		then the pairwise optimal weight satisfies
		\begin{equation}
			\alpha_{i,k,m}^{\star}
			\propto
			\frac{\mu_{i,k,m}}{\sigma_{i,k,m}^{2}}.
		\end{equation}
		Thus, a branch receives more weight when it provides a larger expected target-versus-competitor margin and less weight when that margin is noisy.
	\end{remark}
	
	We next state a complementary single-view recovery condition. Let $y_i$ be the shared ID defined in Equation~\eqref{eq:shared_mapping}, and let branch $r$ retrieve $y_{i,\mathrm{rec}}^{(r)}=\arg\max_k s_{i,k}^{(r)}$ using only its own similarity scores.
	
	\begin{proposition}[Single-view recovery under branch margin]
		\label{prop:single_view_recovery}
		If branch $r$ has a positive one-vs-all margin
		\[
		\gamma_i^{(r)}
		:=
		s_{i,y_i}^{(r)}-\max_{k\neq y_i}s_{i,k}^{(r)}
		>0,
		\]
		then branch $r$ alone recovers the shared ID, i.e., $y_{i,\mathrm{rec}}^{(r)}=y_i$. If the branch scores are perturbed by at most $\epsilon$ in $\ell_\infty$ norm and $\gamma_i^{(r)}>2\epsilon$, the recovered ID is unchanged.
	\end{proposition}
	
	\begin{proof}
		The positive margin condition implies $s_{i,y_i}^{(r)}>s_{i,k}^{(r)}$ for every $k\neq y_i$, so the branch-only argmax is $y_i$. Under a perturbation bounded by $\epsilon$, the score of $y_i$ can decrease by at most $\epsilon$ and any competitor can increase by at most $\epsilon$, so the perturbed margin remains at least $\gamma_i^{(r)}-2\epsilon>0$. The argmax is therefore unchanged.
	\end{proof}
	
	This condition explains how a single temporal view can recover the public ID without requiring all branches to be identical. It also clarifies why the empirical recovery rates in Table~\ref{tab:main_shared_id_match} need not be exactly one: branch-only recovery succeeds where the branch-specific margin is positive and sufficiently stable, while the full shared vote can still use complementary evidence from the other branch.

    \section{Details of Baselines}
    \label{appendix:baselines}

    In this part, we briefly introduce the baselines:

    \noindent\textbf{TimeGAN}~\citep{yoon2019time} is a method that combines Generative Adversarial Networks (GANs) with sequence modeling. It jointly optimizes supervised and unsupervised adversarial losses to learn both the static features and dynamic dependencies of time series data, thereby generating high-quality synthetic time series.
    
    \noindent\textbf{COT-GAN}~\citep{xu2020cot} is a GAN-based method that incorporates causal optimal transport theory. It optimizes the distribution matching between generated and real data to produce high-quality synthetic time series.
    
    \noindent\textbf{TimeVAE}~\citep{desai2021timevae} is a time series generation method based on Variational Autoencoders (VAEs). It enhances model interpretability by incorporating domain-specific temporal patterns (such as polynomial trends and seasonality) and optimizes training efficiency to accurately represent and efficiently synthesize the temporal attributes of the original data.

    \noindent\textbf{DiffWave}~\citep{kong2021diffwave} is an audio synthesis method based on diffusion probabilistic models. The model combines a non-autoregressive structure with bidirectional dilated convolutions to generate high-quality audio in parallel, offering significant advantages in synthesis speed. It also ensures the quality of the generated audio by progressively denoising white noise to produce the target audio waveform.
    
    \noindent\textbf{DiffTime}~\citep{coletta2024constrained} is a time series generation method based on a constrained optimization framework. It combines conditional diffusion models to efficiently generate synthetic time series that meet specific constraints.

    \noindent\textbf{SDformer-ar and SDformer-m}~\citep{chen2024sdformer}
    are two variants of the Similarity-driven Discrete Transformer
    \textbf{SDformer} for time-series generation. SDformer employs
    similarity-driven vector quantization to learn discrete
    time-series representations, followed by Transformer-based
    token distribution modeling. Specifically, SDformer-ar
    adopts autoregressive token generation with sequential
    decoding, whereas SDformer-m employs masked token modeling
    with iterative refinement. Both variants share the same
    discrete tokenization framework but differ in their
    generative strategies.

    \noindent\textbf{MSDformer}~\citep{feng2026msdformer} is a multi-scale extension
    of SDformer, with a particular focus on long-term
    time-series generation. It introduces a coarse-to-fine
    discrete tokenization framework that captures
    multiscale temporal dependencies through coarse
    representations and residual quantization, preserving
    both long-range structures and fine-grained details.
    An autoregressive discrete Transformer then models
    the resulting token sequences to generate time series
    with improved temporal fidelity.

    \noindent\textbf{FlowTS}~\citep{hu2025flowts}
    is a flow-based generative model for time series that
    employs rectified flow to learn approximately straight
    transport trajectories between source and target
    distributions. It incorporates trend-seasonality
    decomposition, attention registers, and rotary positional
    embeddings to capture temporal dependencies and improve
    generation fidelity. An adaptive sampling strategy further
    enhances sampling efficiency, supporting both unconditional
    and conditional time-series generation.

    \section{Model Details}
    \label{appendix:model_details}
    
    This section presents the detailed architectures of
    \method{}. Here, $B$, $d$, and $\ell$ denote the batch
    size, input dimensionality, and sequence length,
    respectively. Stage 1 employs $M$ scale-specific
    encoders with hidden dimension $H$ and branch
    dimensions $\{d_m\}_{m=1}^{M}$, producing a unified
    public-token sequence of length $L=\ell/s$, where
    $s$ denotes the temporal compression ratio and
    $d_c=\sum_m d_m$ is the fused code dimension.
    The encoder, residual block, and shared-ID decoder
    are detailed in Tables~\ref{tab:uniflow_encoder},
    \ref{tab:uniflow_resnet}, and
    \ref{tab:uniflow_decoder}, respectively.
    Stage 2 employs a DiT-based categorical flow generator,
    where $D$, $N_f$, $A$, and $r$ denote the Transformer
    hidden dimension, number of layers, attention heads,
    and anchor rank, respectively.
    Its architecture is summarized in
    Table~\ref{tab:uniflow_stage2}.

    \section{Visualization Methods}
    \label{app:visualization}
    
    To qualitatively assess the distributional fidelity of
    generated time series, we employ two complementary
    visualization techniques: t-distributed Stochastic
    Neighbor Embedding (t-SNE) and Kernel Density Estimation
    (KDE).
    
    \noindent\textbf{t-SNE}~\citep{van2008visualizing}
    projects high-dimensional time-series representations
    into a two-dimensional space by preserving local
    neighborhood relationships. It constructs pairwise
    similarities using adaptive Gaussian kernels and
    optimizes the low-dimensional embedding through
    Kullback--Leibler divergence minimization.
    By jointly visualizing real and generated samples,
    t-SNE provides a qualitative comparison of their
    local distributional structures.
    
    \noindent\textbf{Kernel Density Estimation (KDE)}
    estimates continuous probability densities through
    kernel-based smoothing. Applied to real and generated
    time-series features, KDE enables visual comparison
    of their marginal or joint distributions, highlighting
    differences in density concentration, dispersion,
    and distributional modes.
    
    These two visualization techniques offer complementary perspectives on generative fidelity. Specifically, t-SNE reveals local neighborhood structures, whereas KDE characterizes the distributional properties of temporal features.

    \section{Overview of UniFlow}
    \label{appendix:overview}
    
    \method{} is a unified framework that integrates
    multi-scale representation learning with efficient
    non-autoregressive time-series generation.
    In Stage 1, shared-index voting consolidates
    complementary temporal information into a single
    public-ID sequence, while scale-specific codebooks
    preserve multiscale reconstruction details.
    In Stage 2, geometry-aware categorical flow matching
    combines learned low-rank anchor initialization with
    continuous transport to generate the unified
    public-token representation.
    
    More specifically, \method{} extends SDFlow~\citep{li2026sdflow} toward efficient multi-scale time series generation. Building upon its non-autoregressive categorical flow framework and learned low-rank anchor prior, \method{} introduces shared-index multi-scale tokenization to integrate complementary temporal information into a unified public-ID sequence. This extension enables multiscale generative modeling without increasing the generator-facing sequence length or requiring separate scale-specific generative streams.

    
    
    
    
    
    
    

    \begin{table*}[t]
        \centering
        \small
        \resizebox{0.95\textwidth}{!}{
        \begin{tabular}{@{}cllcc@{}}
            \toprule
            \textbf{Layer}
            & \textbf{Function}
            & \textbf{Description}
            & \textbf{Input Shape}
            & \textbf{Output Shape} \\
            \midrule
    
            1 & Input permutation
            & Channel-first representation
            & $(B,\ell,d)$
            & $(B,d,\ell)$ \\
    
            2 & Patch Conv1D
            & $d\to H$, kernel/stride $p_m$
            & $(B,d,\ell)$
            & $(B,H,\ell_{m,0})$ \\
    
            3 & ReLU
            & Activation
            & $(B,H,\ell_{m,0})$
            & $(B,H,\ell_{m,0})$ \\
    
            4 & ResNet
            & Depth $n_m$, dilation growth $g_m$
            & $(B,H,\ell_{m,0})$
            & $(B,H,\ell_{m,0})$ \\
    
            5 & Downsampling + ResNet
            & Stride-2 Conv1D + ResNet,
            repeated $q_m$ times
            & $(B,H,\ell_{m,j-1})$
            & $(B,H,\ell_{m,j})$ \\
    
            6 & Output Conv1D
            & $H\to d_m$, kernel $3$
            & $(B,H,L)$
            & $(B,d_m,L)$ \\
    
            \bottomrule
        \end{tabular}}
        \caption{Detailed architecture of the Stage 1 multiscale
        encoders. Each branch employs scale-specific patching
        and residual modeling to produce aligned temporal
        representations of length $L$.}
        \label{tab:uniflow_encoder}
    \end{table*}

    \begin{table*}[t]
        \centering
        \small
        \resizebox{0.85\textwidth}{!}{
        \begin{tabular}{@{}cllcc@{}}
            \toprule
            \textbf{Layer}
            & \textbf{Function}
            & \textbf{Description}
            & \textbf{Input Shape}
            & \textbf{Output Shape} \\
            \midrule
    
            1 & ReLU
            & Activation
            & $(B,H,L_r)$
            & $(B,H,L_r)$ \\
    
            2 & Dilated Conv1D
            & $H\to H$, kernel $3$,
            dilation $\delta$, padding $\delta$
            & $(B,H,L_r)$
            & $(B,H,L_r)$ \\
    
            3 & ReLU
            & Activation
            & $(B,H,L_r)$
            & $(B,H,L_r)$ \\
    
            4 & Pointwise Conv1D
            & $H\to H$, kernel $1$
            & $(B,H,L_r)$
            & $(B,H,L_r)$ \\
    
            5 & Residual connection
            & Output = Input + Conv output
            & $(B,H,L_r)$
            & $(B,H,L_r)$ \\
    
            \bottomrule
        \end{tabular}}
        \caption{Detailed architecture of the Stage 1
        residual convolutional block, where $\delta$
        denotes the dilation rate and $L_r$ the
        temporal feature length.}
        \label{tab:uniflow_resnet}
    \end{table*}

    \begin{table*}[t]
        \centering
        \small
        \resizebox{0.98\textwidth}{!}{
        \begin{tabular}{@{}cllcc@{}}
            \toprule
            \textbf{Layer}
            & \textbf{Function}
            & \textbf{Description}
            & \textbf{Input Shape}
            & \textbf{Output Shape} \\
            \midrule
    
            1 & Shared-index voting
            & Weighted cosine similarity,
            vocabulary size $K$
            & $\{(B,d_m,L)\}_{m=1}^{M}$
            & $(B,L)$ \\
    
            2 & Codebook retrieval
            & Shared-ID lookup in
            scale-specific codebooks
            & $(B,L)$
            & $\{(B,L,d_m)\}_{m=1}^{M}$ \\
    
            3 & Feature fusion
            & Concatenate private embeddings,
            $d_c=\sum_m d_m$
            & $\{(B,L,d_m)\}_{m=1}^{M}$
            & $(B,d_c,L)$ \\
    
            4 & Input Conv1D
            & $d_c\to H$, kernel $3$
            & $(B,d_c,L)$
            & $(B,H,L)$ \\
    
            5 & ReLU
            & Activation
            & $(B,H,L)$
            & $(B,H,L)$ \\
    
            6 & ResNet
            & Depth $n_{\mathrm{dec}}$,
            reverse dilation
            & $(B,H,2^{j-1}L)$
            & $(B,H,2^{j-1}L)$ \\
    
            7 & Upsample + Conv1D
            & Nearest-neighbor upsampling
            $\times 2$ + Conv1D
            & $(B,H,2^{j-1}L)$
            & $(B,H,2^jL)$ \\
    
            $\cdots$ & $\cdots$
            & Repeat layers 6--7
            for $T_d$ operations
            & $\cdots$ & $\cdots$ \\
    
            8 & Output projection
            & Conv1D + ReLU + Conv1D,
            $H\to d$
            & $(B,H,2^{T_d}L)$
            & $(B,d,2^{T_d}L)$ \\
    
            9 & Output alignment
            & Match length $\ell$ and permute
            & $(B,d,2^{T_d}L)$
            & $(B,\ell,d)$ \\
    
            \bottomrule
        \end{tabular}}
        \caption{Detailed architecture of Stage 1 shared-ID
        assignment and reconstruction. Shared public IDs
        retrieve scale-specific codebook embeddings, which
        are fused for time-series reconstruction.}
        \label{tab:uniflow_decoder}
    \end{table*}

    \begin{table*}[t]
        \centering
        \small
        \resizebox{0.98\textwidth}{!}{
        \begin{tabular}{@{}cllcc@{}}
            \toprule
            \textbf{Layer}
            & \textbf{Function}
            & \textbf{Description}
            & \textbf{Input Shape}
            & \textbf{Output Shape} \\
            \midrule
    
            1 & Low-rank anchor
            & Projection through
            $V\in\mathbb{R}^{r\times Ld_c}$
            & $(B,r)$
            & $(B,L,d_c)$ \\
    
            2 & Flow interpolation
            & $Z_t=(1-t)Z_0+tZ_1$
            & $Z_0,Z_1:(B,L,d_c)$
            & $(B,L,d_c)$ \\
    
            3 & Global token
            & Prepend one learned token
            & $(B,L,d_c)$
            & $(B,L+1,d_c)$ \\
    
            4 & Input projection
            & Linear $d_c\to D$
            + positional encoding
            & $(B,L+1,d_c)$
            & $(B,L+1,D)$ \\
    
            5 & Time embedding
            & Linear $1\to4D$ + SiLU
            + Linear $4D\to4D$
            & $(B,1)$
            & $(B,4D)$ \\
    
            6 & DiT blocks
            & AdaLN + MHSA ($A$ heads)
            + FFN ($4D$), repeated $N_f$ times
            & $(B,L+1,D)$
            & $(B,L+1,D)$ \\
    
            7 & Categorical head
            & LayerNorm + Linear $D\to K$,
            remove global-token output
            & $(B,L+1,D)$
            & $(B,L,K)$ \\
    
            8 & Endpoint estimation
            & Softmax + codebook expectation
            & $(B,L,K)$
            & $(B,L,d_c)$ \\
    
            9 & ODE transport
            & Parallel velocity-based integration
            & $(B,L,d_c)$
            & $(B,L,d_c)$ \\
    
            10 & Public-ID readout
            & Normalize + nearest-code assignment
            & $(B,L,d_c)$
            & $(B,L)$ \\
    
            \bottomrule
        \end{tabular}}
        \caption{Detailed architecture of the Stage 2
        non-autoregressive categorical flow generator.
        The DiT predicts categorical endpoint distributions
        for joint transport of all public-token positions.}
        \label{tab:uniflow_stage2}
    \end{table*}

    
    
    
    
    
    
    
    
    
    
    

    \begin{figure*}[t]
        \centering
        \includegraphics[width=0.95\textwidth]{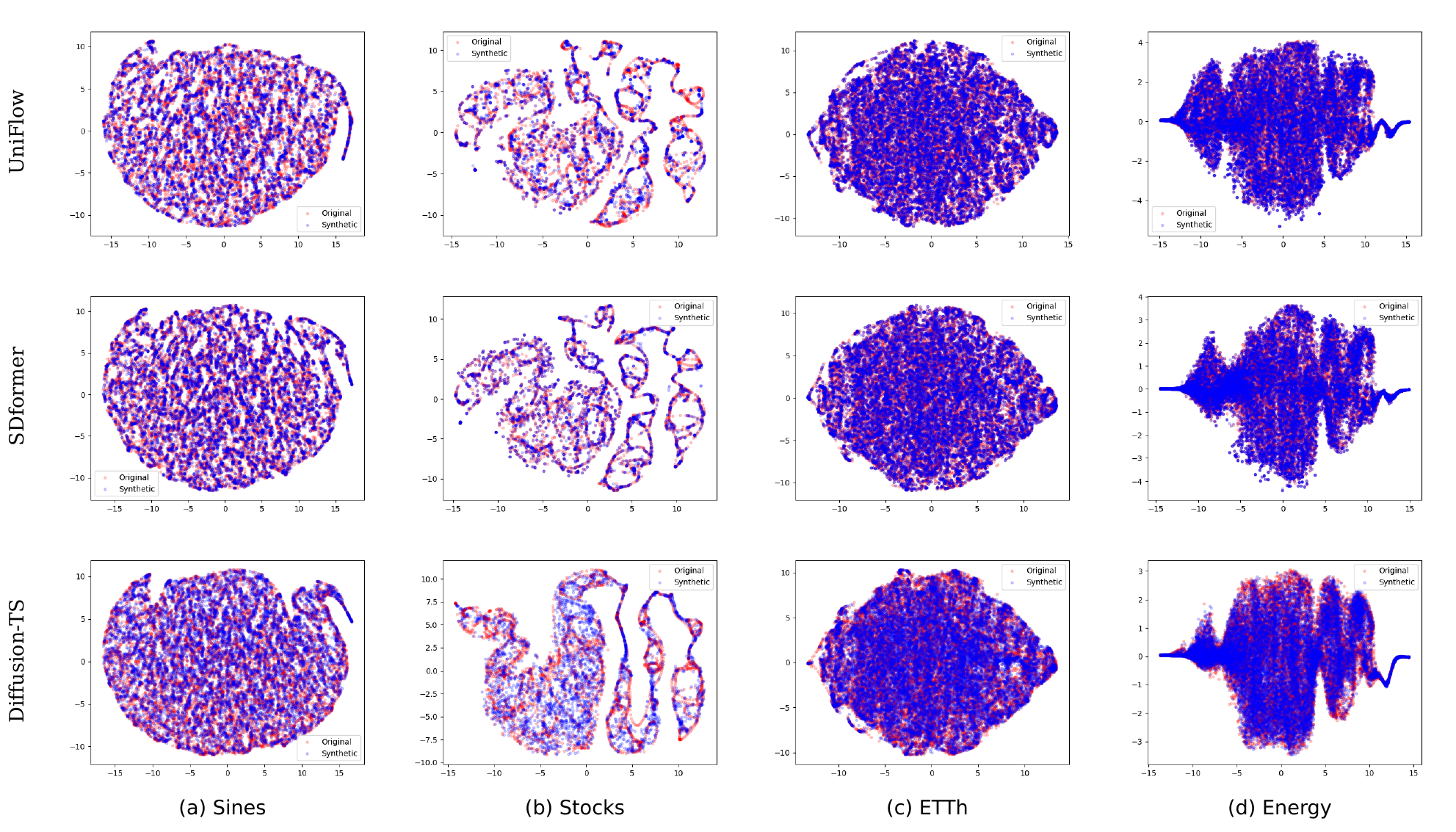}
        \caption{t-SNE visualizations of the time series synthesized by UniFlow, SDformer and Diffusion-TS.}
        \label{fig:app_tsne}
    \end{figure*}

    \begin{figure*}[t]
        \centering
        \includegraphics[width=0.95\textwidth]{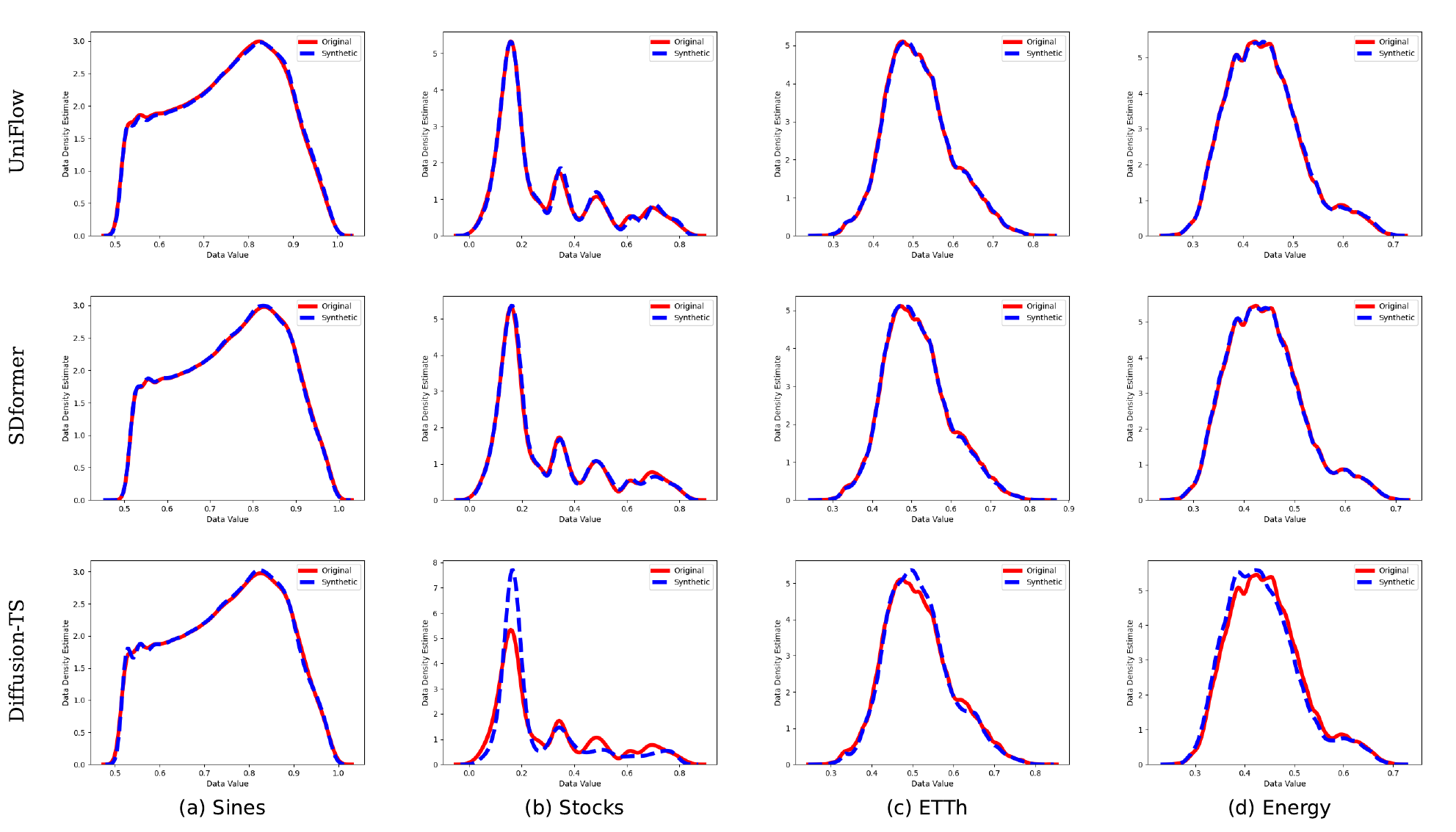}
        \caption{Kernel density estimation visualizations of the time series synthesized by UniFlow, SDformer and Diffusion-TS.}
        \label{fig:app_kernel}
    \end{figure*}

\end{document}